\documentclass{article}
\usepackage[12pt]{extsizes}
\usepackage[utf8]{inputenc}
\usepackage{amsmath}
\usepackage{amssymb}
\usepackage{amsthm}
\usepackage{natbib}
\usepackage{graphicx}
\usepackage{lipsum}
\usepackage{enumitem}

\usepackage[dvipsnames]{xcolor}
\usepackage[hidelinks]{hyperref}

\definecolor{myurlcolor}{HTML}{123463}
\definecolor{tdc_color}{RGB}{10,128,122}
\hypersetup{
    colorlinks=true,
    linkcolor=myurlcolor,
    filecolor=myurlcolor,      
    urlcolor=myurlcolor,
    citecolor=myurlcolor
}

\usepackage{authblk}

\usepackage{geometry}
\usepackage{tocloft}

\usepackage{mathtools}
\usepackage{amsfonts}
\usepackage{relsize}
\usepackage{multirow}
\usepackage{booktabs}
\usepackage{tabularx}
\usepackage{bbding}
\usepackage{listings}

\usepackage{xspace}
\usepackage[ruled,vlined]{algorithm2e}

\usepackage{multirow}

\newtheorem*{problem*}{Problem}

\definecolor{light-gray}{gray}{0.96}

\newcommand{\xhdr}[1]{\vspace{2mm}\noindent{{\bf #1.}}}
\newcommand{\hide}[1]{}

\usepackage[p,osf]{cochineal}
\usepackage[scale=.95,type1]{cabin}
\usepackage[cochineal,bigdelims,cmintegrals,vvarbb]{newtxmath}
\usepackage[zerostyle=c,scaled=.94]{newtxtt}
\usepackage[cal=boondoxo]{mathalfa}
\usepackage{microtype}
\usepackage{multirow}
\usepackage{adjustbox}

\usepackage{etoolbox}
\apptocmd{\thebibliography}{\raggedright}{}{}

\usepackage{lmodern}

\usepackage{dirtree}

\makeatletter
\patchcmd{\@maketitle}{\LARGE \@title}{\fontsize{25}{30}\selectfont\@title}{}{}
\makeatother

\usepackage{amsmath,amsfonts,bm}

\def\eqref#1{equation~\ref{#1}}

\DeclareMathAlphabet{\mathsfit}{\encodingdefault}{\sfdefault}{m}{sl}
\SetMathAlphabet{\mathsfit}{bold}{\encodingdefault}{\sfdefault}{bx}{n}

\title{Machine Learning Applications for Therapeutic Tasks with Genomics Data }

\author[1,2]{Kexin Huang*}
\author[2]{Cao Xiao*}
\author[2]{Lucas M. Glass}
\author[3]{Cathy W. Critchlow}
\author[4]{\\Greg Gibson}
\author[5]{Jimeng Sun}

\affil[1]{Health Data Science Program, Harvard University, Boston, MA 02120}
\affil[2]{Analytics Center of Excellence, IQVIA, Cambridge, MA 02139}
\affil[3]{Center for Observational Research, Amgen, Thousand Oaks, CA 91320}
\affil[4]{Center for Integrative Genomics, Georgia Institute of Technology, Atlanta, GA 30332}
\affil[5]{Department of Computer Science and Carle's Illinois College of Medicine, University of Illinois at Urbana-Champaign, Urbana, IL 61820}

\date{}

\begin{document}
\maketitle

\begin{abstract}
    The genome contains instructions for building the function and structure, and guiding the evolution of molecules and organisms. Recent high-throughput techniques allow the generation of a vast amount of genomics data. However, the path of transforming genomics data into tangible therapeutics is filled with obstacles. We observe that genomics data alone are insufficient but rather require investigation of its interplay with data such as compounds, proteins, electronic health records, images, texts, etc. To make sense of these complex data, machine learning techniques are often utilized for identifying patterns and drawing insights from data. In this review, we study twenty-two genomics applications of machine learning that can enable faster and more efficacious therapeutic development, from discovering novel targets, personalized medicine, developing gene-editing tools all the way to clinical trials and post-market studies. Challenges remain, including technical issues such as learning under different contexts given specific low resource constraints, and practical issues such as mistrust of models, privacy, and fairness.
\end{abstract}




\section{Introduction}

Genomics studies the function, structure, evolution, mapping, and editing of genomes~\citep{hieter1997functional}. The genome contains chapters of instructions for building various types of molecules and organisms. Probing genomes allows us to understand a biological phenomenon, such as identifying the roles that the genome play in diseases. A deep understanding of genomics has led to a vast array of successful therapeutics to cure a wide range of diseases, both complex and rare~\citep{wong2004monoamines,chin2011cancer}. It also allows us to prescribe more precise treatments~\citep{hamburg2010path}, or seek more effective therapeutics strategies such as genome editing~\citep{makarova2011evolution}.

Recent advances in high-throughput technologies have led to an outpouring of large-scale genomics data~\citep{reuter2015high,heath2021nci}. However, the bottlenecks along the path of transforming genomics data into tangible therapeutics are innumerable. For instance, diseases are driven by multifaceted mechanisms, so to pinpoint the right disease target requires knowledge about the entire suite of biological processes, including gene regulation by non-coding regions~\citep{rinn2012genome}, DNA methylation status~\citep{singal1999dna}, and RNA splicing~\citep{rogers1980mechanism}; personalized treatment requires accurate characterization of disease sub-types, and the compound's sensitivity to various genomics profiles~\citep{hamburg2010path}; gene-editing tools require an understanding of the interplay between guide RNA and the whole-genome to avoid off-target effects~\citep{fu2013high}; monitoring therapeutics efficacy and safety after approval requires the mining of gene-drug-disease relations in the EHR and literature~\citep{corrigan2018real}. We argue that genomics data alone are insufficient to ensure clinical implementation, but it requires integration of a diverse set of data types, from multi-omics, compounds, proteins, cellular image, electronic health records (EHR), and scientific literature. This heterogeneity and scale of data enable application of sophisticated computational methods such as machine learning (ML). 

Over the years, ML has profoundly impacted many application domains, such as computer vision~\citep{krizhevsky2012imagenet}, natural language processing~\citep{devlin2018bert}, and complex systems ~\citep{silver2016mastering}. ML has changed computational modeling from expert-curated features to automated feature construction. It can learn useful and novel patterns from data, often not found by experts, to improve prediction performance on various tasks. This ability is much-needed in genomics and therapeutics as our understanding of human biology is vastly incomplete. Uncovering these patterns can also lead to the discovery of novel biological insights. Also, therapeutic discovery consists of large-scale resource-intensive experiments, which limit the scope of experiment and many potent candidates are therefore missed. Using accurate prediction by ML can drastically scale up and facilitate the experiments, catching or generating novel therapeutics candidates.

Interests in ML for genomics through the lens of therapeutic development have also grown for two reasons. First, for pharmaceutical and biomedical researchers, ML models have undergone proof-of-concept stages in yielding astounding performance often of previously infeasible tasks~\citep{stokes2020deep,senior2020improved}. Second, for ML scientists, large/complex data and hard/impactful problems present exciting opportunities for innovation. 

This survey summarizes recent ML applications related to genomics in therapeutic development and describes associated challenges and opportunities. Several reviews of ML for genomics have been published~\citep{leung2015machine,eraslan2019deep,zou2019primer}. Most of these previous works focused on studying genomics for biological applications, whereas we study them in the context of bringing genomics discovery to therapeutic implementations. We identify twenty-two “ML for therapeutics” tasks with genomics data, ranging across the entire therapeutic pipeline, which were not covered in previous surveys. Moreover, most of the previous reviews focused on DNA sequences, while we go beyond DNA sequences and study a wide range of interactions among DNA sequences, compounds, proteins, multi-omics, and EHR data. 

In this survey, we organize ML applications into four therapeutic pipelines: (1) target discovery: basic biomedical research to discover novel disease targets to enable therapeutics; (2) therapeutic discovery: large-scale screening designed to identify potent and safe therapeutics; (3) clinical study: evaluating the efficacy and safety of the therapeutics in vitro, in vivo, and through clinical trials; and (4) post-market study: monitoring the safety and efficacy of marketed therapeutics and identifying novel indications. We also formulate these tasks and data modalities in ML languages, which can help ML researchers with limited domain background to understand those tasks. In summary, this survey presents a unique perspective on the intersection of machine learning, genomics, and therapeutic development. 

The survey is organized as follows. In Section~\ref{sec:primer}, we provide a brief primer on genomics-related data. We also review popular machine learning model for each data type. Next, in Sections~\ref{sec:target}-\ref{sec:post-market}, we discuss ML applications in genomics across the therapeutics development pipeline. Each section describes a phase in the therapeutics pipeline and contains several ML applications and ML models and formulations. Lastly, in Section~\ref{sec:challenge}, we identify seven open challenges that present numerous opportunities for ML model development and also novel applications.

\begin{figure}[t]
    \centering
    \includegraphics[width=0.95\textwidth]{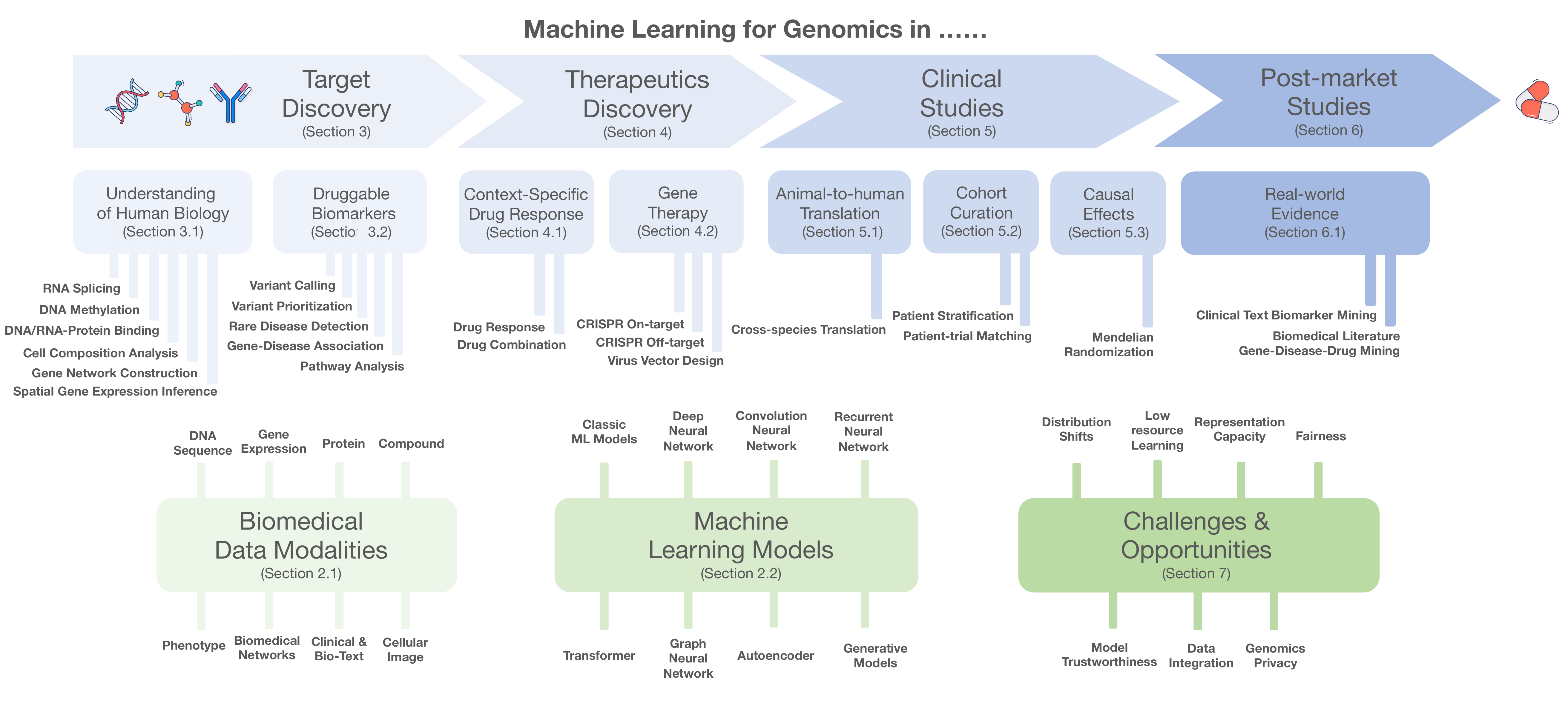}
    \caption{\textbf{Organization and coverage of this survey}. Our survey covers a wide range of important ML applications in genomics across the therapeutics pipelines (Section~\ref{sec:target}-\ref{sec:post-market}). In addition, we provide a primer on biomedical data modalities and machine learning models (Section~\ref{sec:primer}). At last, we identify seven challenges filled with opportunities (Section~\ref{sec:challenge}).}
    \label{fig:summary}
\end{figure}
\section{A Primer on Genomics Data and Machine Learning Models} \label{sec:primer}

With advances in high-throughput technologies and data management systems, we now have vast and heterogeneous datasets in the field of biomedicine. This section introduces the basic genomics-related data types and their machine learning representation and provides a primer on popular machine learning methods applied to these data. 

\subsection{Genomics-related biomedical data} \label{sec:data}

\begin{figure}[t]
    \centering
    \includegraphics[width=0.9\textwidth]{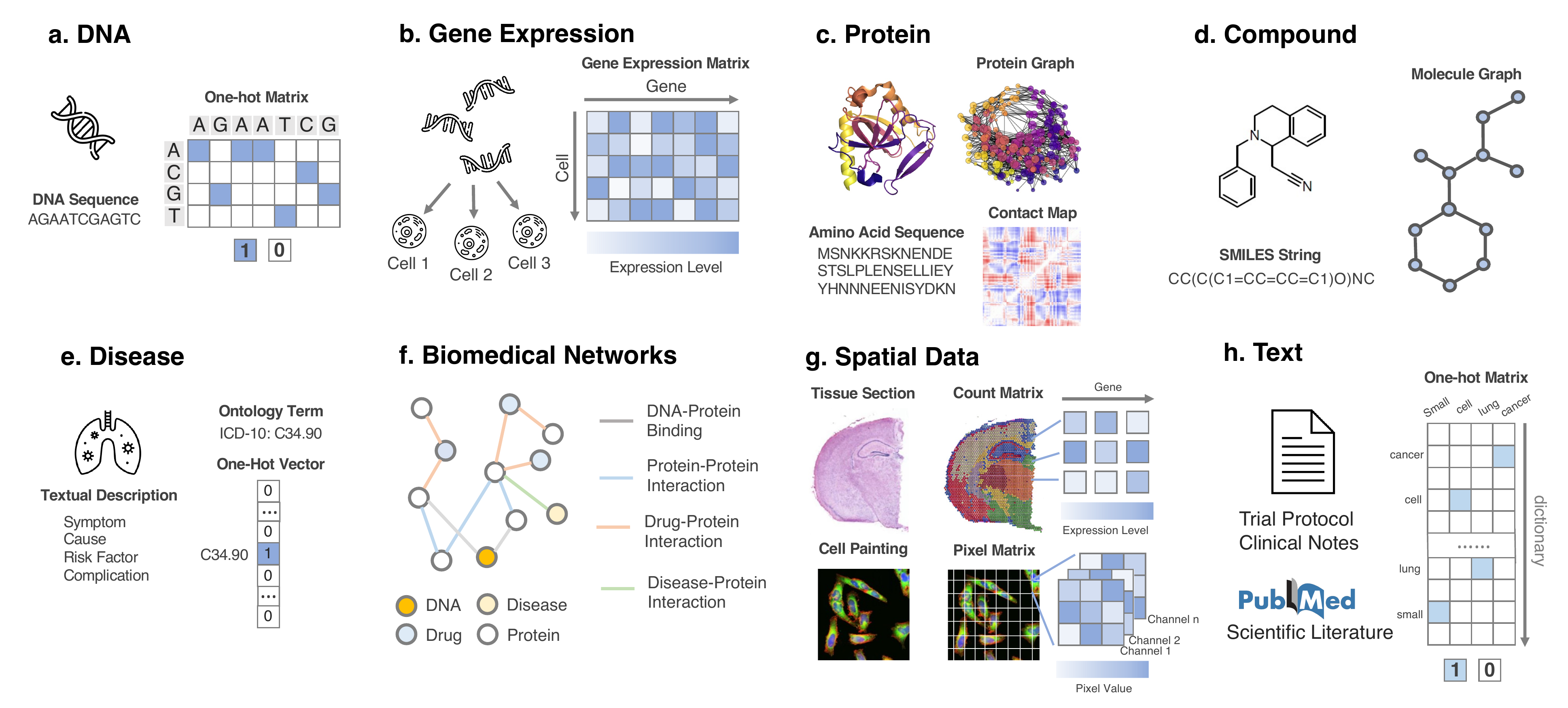}
    \caption{\textbf{Therapeutics data modalities and their machine learning representation.} Detailed descriptions of each modality can be found in Section~\ref{sec:data}. \textbf{a.} DNA sequences can be represented as a matrix where each position is a one-hot vector corresponding to A, C, G, T. \textbf{b.} Gene expressions are a matrix of real value, where each entry is the expression level of a gene in a context such as a cell. \textbf{c.} Proteins can be represented in amino acid strings, a protein graph, and a contact map where each entry is the connection between two amino acids. \textbf{d.} Compounds can be represented as a molecular graph or a string of chemical tokens, which are a depth-first traversal of the graph. \textbf{e.} Diseases are usually described by textual descriptions and also symbols in the disease ontology. \textbf{f.} Networks connect various biomedical entities with diverse relations. They can be represented as a heterogeneous graph. \textbf{g.} Spatial data are usually depicted as a 3D array, where 2 dimensions describe the physical position of the entity and the 3rd dimension corresponds to colors (in cell painting) or genes (in spatial transcriptomics). \textbf{h.} Texts are typically represented as a one-hot matrix where each token corresponds to its index in a static dictionary.  Credits: The protein image is adapted from \cite{gaudelet2020utilising}; the spatial transcriptomics image is adapted from 10x Genomics; the cell painting image is from Charles River Laboratories.}
    \label{fig:data}
\end{figure}

\xhdr{DNAs} The human genome can be thought of as the instructions for building functional individuals. DNA sequences encode these instructions. Like a computer, where we build a program based on 0/1 bit, the basic DNA sequence units are called nucleotides (A, C, G, and T). Given a list of nucleotides, a cell can build a diverse range of functional entities (programs). There are approximately 3 billion base pairs for the human genome, and more than 99.9\% are identical between individuals. If a subset of the population has different nucleotides in a genome position than the majority, this position is called a variant. This single nucleotide variant is often called a single nucleotide polymorphism (SNP). While most variants are not harmful (they are said to be functionally neutral), many correspond to the potential driver for phenotypes, including diseases. 

\textit{Machine learning representations:} A DNA sequence is a list of ACGT tokens of length $N$. It is typically represented in three ways: (1) a string $\{A, C, G, T\}^N$; (2) a two dimensional matrix $\mathbf{W} \in \mathbb{R}^{4 \times N}$, where the $i$-th column $\mathbf{W}_i$ corresponds to the $i$-th nucleotide and is an one-hot encoding vector of length 4, where A, C, T and G are encoded as [1,0,0,0], [0,1,0,0], [0,0,1,0], and [0,0,0,1], respectively; or (3) a vector of $\{0, 1\}^N$, where 0 means it is not a variant, and 1 a variant. Example illustration in Figure \ref{fig:data}a.

\xhdr{Gene expression/transcripts} In a cell, the DNA sequence of each gene is transcribed into messenger RNA (mRNA) transcripts. While most cells share the same genome, the individual genes are expressed at very different levels across cells and tissue types and given different interventions and environments. These expression levels can be measured by the count of mRNA transcripts. Given a disease, we can compare the gene expression in people with the disease with expression to people in healthy cohorts (without the disease of interest) and associate various genes with the underlying biological processes in this disease. With the advance of single-cell RNA sequencing (scRNA-seq) technology, we can now obtain gene expression for the different types of cells that make up a tissue. The availability of transcriptomes of tens of thousands of cells creates new opportunities for understanding interactions among cell types and the impact of heterogeneity. 

\textit{Machine learning representations:} Gene expressions/transcripts are counts of mRNA. For a scRNA-seq experiment, given $M$ cells with $N$ genes, we can obtain a gene expression matrix $\mathbf{W} \in \mathbb{Z}^{M \times N}$, where each entry $\mathbf{W}_{i,j}$ corresponds to the transcript counts of gene $j$ for cell $i$. Example illustration in Figure \ref{fig:data}b.

\xhdr{Proteins} Most of the genes encoded in the DNA provide instructions to build a diverse set of proteins, which perform a vast array of functions. For example, transcription factors are proteins that bind to the DNA/RNA sequence, and regulate their expression in different conditions. A protein is a macro-molecule and is represented by a sequence of 20 standard amino acids or residues, where each amino acid is a simple compound. Based on this sequence code, it naturally folds into a 3D structure, which determines its function. As the functional units, proteins present a large class of therapeutic targets. Many drugs are designed to inhibit/promote proteins in the disease pathways. Proteins can also be used as therapeutics such as antibodies and peptides. 

\textit{Machine learning representations:} Proteins have diverse forms. For a protein with $N$ amino acids, it can be represented in the following formats: (1) a string $\{A, R, N, D, ...\}^N$ of amino acid sequence tokens; (2) a contact map matrix $\mathbf{W} \in \mathbb{R}^{N \times N}$ where $\mathbf{W}_{i,j}$ is the physical distance between $i$-th and $j$-th amino acids; (3) a protein graph $G$ with nodes corresponding to amino acids, where nodes are connected based on rules such as a physical distance threshold or k-nearest neighbors; (4) a protein 3D grid with three-dimensional discretized tensor, where each grid point $(x, y, z)$ corresponds to amino acids in the 3D space. Example illustration in Figure \ref{fig:data}c.

\xhdr{Compounds} Compounds are molecules that are composed of atoms connected by chemical bonds. They can interact with proteins and drive important biological processes. In their natural form, compounds have a 3D structure. Small-molecule compounds are the major class of therapeutics. 

\textit{Machine learning representations:} A compound is usually represented as (1) a SMILES string where it is a depth traversal order of the molecule graph; (2) a molecular graph $G$ where each node is an atom and edges are the bonds. Example illustration is in Figure \ref{fig:data}d.

\xhdr{Diseases} A disease is an abnormal condition that affects the function and/or modifies the structure of an organism. It is derived from both genotypes and environmental factors, with intricate mechanisms driven by biological processes. They are observable and can be described by certain symptoms.

\textit{Machine learning representations:} Diseases are represented by (1) symbols such as disease ontology; (2) text description of the specific disease.  Example illustration is in Figure \ref{fig:data}e.

\xhdr{Biomedical networks} Biological processes are not driven by individual units but consist of numerous interactions among various types of entities such as cell signaling pathways, protein-protein interactions, and gene regulation. These interactions can be characterized by biomedical networks, where they provide a systems view toward biological phenomena. 

\textit{Machine learning representations:} Biomedical networks are represented as graphs, where each node is a biomedical entity, and an edge corresponds to relations among them.  Example illustration is in Figure \ref{fig:data}f.

\xhdr{Spatial data} With the advance of microscopes and fluorescent probes, we can visualize cell dynamics through cellular images. By imaging cells under various conditions such as drug treatment, they allow us to identify the effect of conditions at a cellular level. Furthermore, spatial genomic sequencing techniques now allow us to visualize and understand the gene expression for cellular processes in the tissue environment. 

\textit{Machine learning representations:} Cellular image or spatial transcriptomics can be represented as a matrix of size $M \times N$, where $M, N$ is the width and height of the data or number of pixels/transcripts along this dimension, and each entry corresponds to the pixel of the image or the transcript count in the case of spatial transcriptomics. Additional channels (a separate matrix of size $M \times N$) to encode for information such as colors or various genes for spatial transcriptomics. After aggregation, the spatial data can be represented as a tensor of size $M \times N \times H$, where $H$ is the number of channels. Example illustration in Figure \ref{fig:data}g.

\xhdr{Texts} The first important example of text encountered in therapeutics development include clinical trial design protocols, where texts describe inclusion and exclusion criteria for trial participation, often as a function of genome markers. For example, in a trial to study Gefitinib for EGFR-mutant Non-Small Cell Lung Cancer, one of the trial eligibility criteria would be "An EGFR sensitizing mutation must be detected in tumor tissue"~\citep{trial}. A second type of clinical text is clinical notes documented in electronic health records, containing valuable information for post-market research on treatments. 

\textit{Machine learning representations:} Clinical texts are similar to texts in common natural language processing. The standard way to represent them is a matrix of size $M \times N$, where $M$ is the number of total vocabularies and $N$ is the number of tokens in the texts. Each column is a one-hot encoding for the corresponding token. Example illustration is in Figure \ref{fig:data}h.

\subsection{Machine Learning Methods for Biomedical Data} \label{sec:methods}

Machine learning models learn patterns from data and leverage these patterns to make accurate predictions. Numerous ML models have been proposed to tackle different challenges. In this section, we briefly introduce the main mechanisms of popular ML models used to analyze genomic data. 

\begin{figure}
    \centering
    \includegraphics[width=0.9\textwidth]{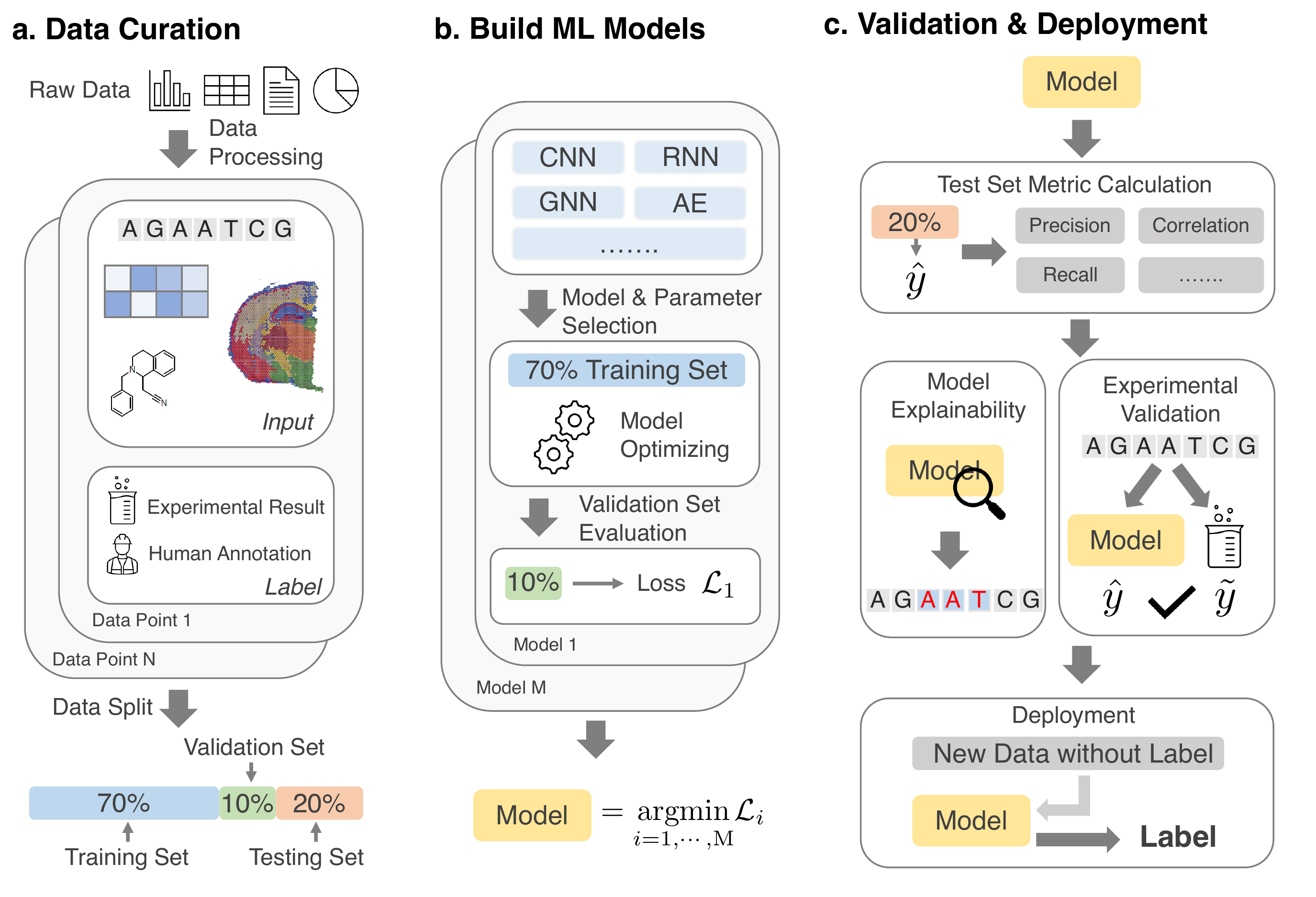}
    \caption{\textbf{Machine learning for genomics workflow.} \textbf{a.} The first step is to curate a machine learning dataset. Raw data are extracted from databases of various sources, and they are processed into data points. Each data point corresponds to an input of a series of biomedical entities and a label from annotation or experimental result. These data points constitute a dataset, and they are split into three sets. The training set is for the ML model to learn and identify useful and generalizable patterns. The validation set is for model selection and parameter tuning. The testing set is for the evaluation of the final model. The data split could be constructed in a way to reflect real-world challenges. \textbf{b.} Various ML models can be trained using the training set and tuned based on  a quantified metric on the validation set such as loss $\mathcal{L}$ that measures how good this model predicts the output given the input. Lastly, we select the optimal model given the lowest loss. \textbf{c.} The optimal model can then predict on the test set, where various evaluation metrics are used to measure how good is the model on new unseen data points. Models can also be probed with explainability methods to identify biological insights captured by the model. Experimental validation is also common to ensure the model can approximate wet-lab experiment results. Finally, the model can be deployed to make predictions on new data without labels. The prediction becomes a proxy for the label from downstream tasks of interest. }
    \label{fig:ml_workflow}
\end{figure}

\xhdr{Preliminary} A typical ML model for genomics usage is as follows: given an input of a set of data points, where each data point consists of input features and a ground truth biological label, a machine learning model aims to learn a mapping from input to a label based on the observed data points, which are also called training data. This setting of predicting by leveraging known supervised labels is also called supervised learning. The size of the training data is called the sample size. ML models are data-hungry and usually need a large sample size to perform well. 

The input features can be DNA sequences, compound graphs, or clinical texts, depending on the task at hand. The ground truth label is usually obtained via biological experiments. The ground truth also presents the goal for an ML model to achieve. Thus, the quality of the ground truth directly affects ML model performance, highlighting the necessity of label curation. There are various forms of ground truth labels. If the labels are continuous (e.g., binding scores), the learning problem is a {\it regression} problem. And if the labels are discrete variables (e.g., the occurrence of interaction), the problem is a {\it classification} problem. Models focusing on predicting the labels of the data are called {\it discriminative models}. Besides making predictions, ML models can also generate new data points by modeling the statistical distribution of data samples. Models following this procedure are called {\it generative models}. 

When labels are not available, an ML model can still identify the underlying patterns within the unlabeled data points. This problem setting is called {\it unsupervised learning}, where models discover patterns or clusters (e.g., cell types) by modeling the relations among data points.
{\it Self-supervised learning} uses supervised learning methods for handling unlabeled data. It creatively produces labels from the unlabeled data (e.g., masking out a motif and use the surrounding context to predict the motif)~\citep{devlin2018bert,hu2019strategies}. 

In many biological cases, ground truth labels are scarce, where few-shot learning can be considered. {\it Few-shot learning} assumes only a few labeled data points but many unlabeled data points. Another strategy is called {\it meta-learning}, which aims to learn from a set of related tasks to form the ability to learn quickly and accurately on an unseen task. 

If a model integrates multiple data modalities (e.g., DNA sequence plus compound structure), it is called {\it multimodal learning}. When a model predicts multiple labels (e.g., multiple target endpoints), it is called {\it multi-task learning}. 

\begin{figure}[t]
    \centering
    \includegraphics[width = 0.85\textwidth]{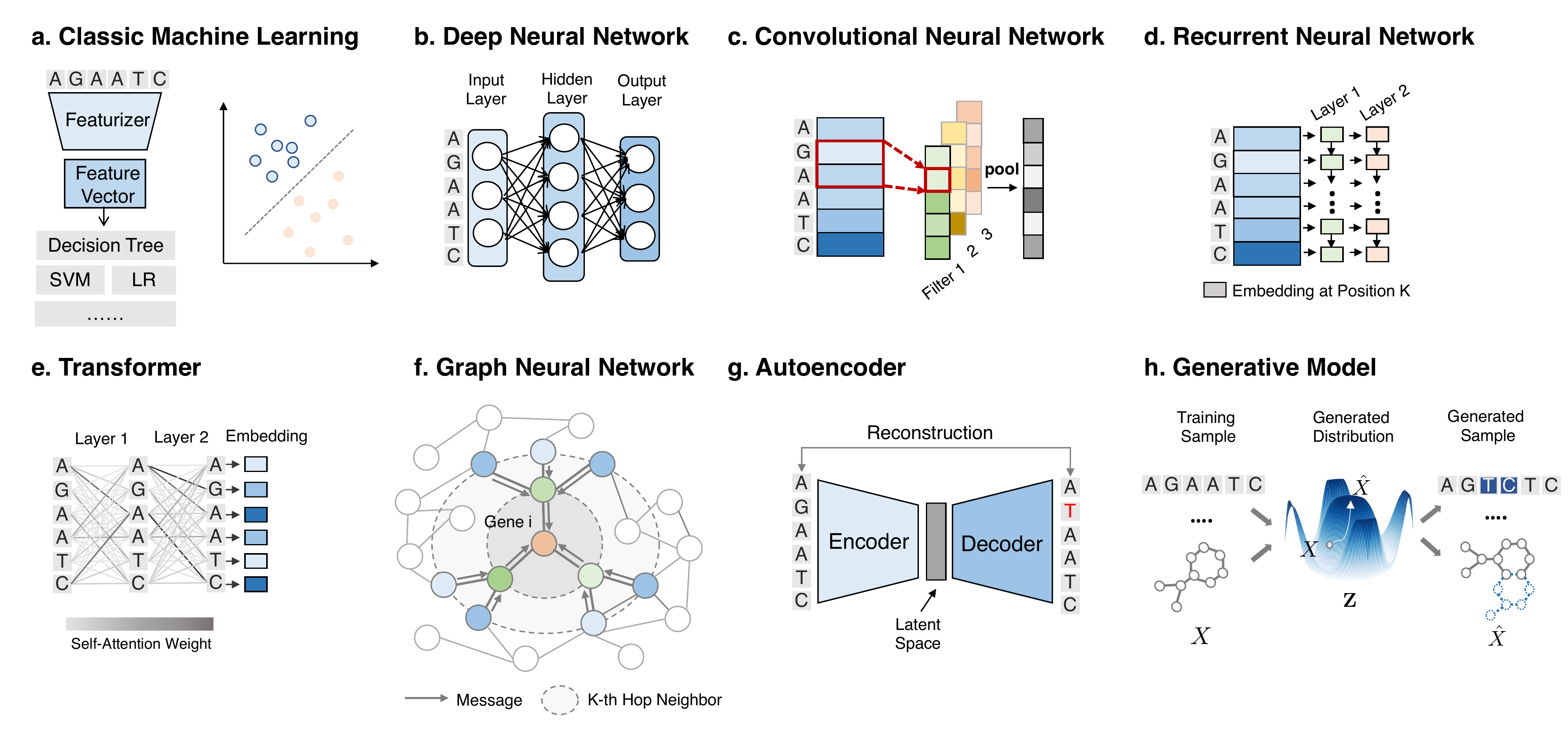}
    \caption{\textbf{Machine learning models illustrations.} Details about each model can be found in Section~\ref{sec:methods}. \textbf{a.} Classic machine learning models featurize raw data and apply various models (mostly linear) to classify (e.g., binary output) or regress (e.g., real value output). \textbf{b.} Deep Neural Networks map input features to embeddings through a stack of non-linear weight multiplication layers. \textbf{c.} Convolutional Neural Networks apply many local filters to extract local patterns and aggregate local signals through pooling. \textbf{d.} Recurrent Neural Networks generate embeddings for each token in the sequence based on the previous tokens. \textbf{e.} Transformers apply a stack of self-attention layers that assign a weight for each pair of input tokens. \textbf{f.} Graph Neural Networks aggregate information from the local neighborhood to update the node embedding. \textbf{g.} Autoencoders reconstruct the input from an encoded compact latent space. \textbf{h.} Generative models generate novel biomedical entities with more desirable properties. }
    \label{fig:models}
\end{figure}

\xhdr{Classic Machine Learning Models} Traditional ML usually requires a transformation of input to tabular real-valued data, where each data point corresponds to a feature vector. In our context, these are predefined features such as the SNP vector, polygenic risk scores, and chemical fingerprints. These tabular data can then be fed into a wide range of supervised models, such as linear/logistic regression, decision trees, random forest, support vector machine, and naive Bayes~\citep{mitchell1997machine}. They work well when the features are well defined. A multilayer perceptron~\citep{rosenblatt1961principles} (MLP) consists of at least three layers of neurons, where each layer is fed into a nonlinear activation function to capture these patterns. When the number of layers is large, it is called a deep neural network (DNN). 

\textit{Suitable biomedical data:} any real-value feature vectors built upon biomedical entities such as SNP profile and chemical fingerprints. 

\xhdr{Convolution Neural Network (CNN)} CNNs represent a class of DNNs widely applied for image classification, natural language processing, and signal processing such as speech recognition~\citep{lecun1995convolutional}.  A CNN model has a series of convolution filters, which allow it to identify local patterns in the data (e.g., edges, shapes for images). Such networks can automatically extract hierarchical patterns in data. The weight of each filter reveals patterns (such as conserved motifs). 

\textit{Suitable biomedical data:} short DNA sequence, compound SMILES strings, gene expression profile, cellular images. 

\xhdr{Recurrent Neural Network (RNN)}
An RNN is designed to model sequential data, such as time series, event sequences, and natural language text~\citep{de2015survey}.  The RNN model is sequentially applied to a sequence. The input at each step includes the current observation and the previous hidden state. RNN is natural to model variable-length sequences. There are two widely used variants of RNNs: long short-term memory (LSTM)~\citep{hochreiter1997long} and gated recurrent units (GRU)~\citep{cho2014properties}.

\textit{Suitable biomedical data:} DNA sequence, protein sequence, texts.

\xhdr{Transformer} Transformers~\citep{vaswani2017attention} are a recent class of neural networks that leverage self-attentions: assigning a score of interaction among every pair of input features (e.g., a pair of DNA nucleotides). By stacking these self-attention units, the model can capture more expressive and complicated interactions. Transformers have shown superior performances on sequence data, such as natural language processing. They have also been successfully adapted for state-of-the-art performances on proteins~\citep{Rivese2016239118} and compounds~\citep{huangmoltrans}. 

\textit{Suitable biomedical data:} DNA sequence, protein sequence, texts, image.

\xhdr{Graph Neural Networks (GNN)} Graphs are universal representations of complex relations in many real-world objects. In biomedicine, graphs can represent knowledge graphs, molecules, protein-protein interaction networks, and medical ontologies. However, graphs do not follow rigid data structures as in sequences and images. GNNs are a class of models that convert graph structures into embedding vectors (i.e., node representation or graph representation vectors)~\citep{kipf2016semi}. In particular, GNNs generalize the concept of convolution operations to graphs by iteratively passing and aggregating messages from neighboring nodes. The resulting embedding vectors capture the node attributes and the network structures. 

\textit{Suitable biomedical data:} biomedical networks, compound/protein graphs, similarity network.

\xhdr{Autoencoders (AE)} Autoencoders are an unsupervised method in deep learning.  Autoencoders map the input data into a latent embedding (encoder) and then reconstruct the input from the latent embedding (decoder)~\citep{kramer1991nonlinear}. Their objective is to reconstruct the input from a low-dimensional latent space, thus allowing the latent representation to focus on essential properties of the data. Both encoders and decoders are neural networks. AE can be considered as a nonlinear analog to principal component analysis (PCA). The generated latent representation capture patterns in the input data and can thus be used to do unsupervised learning tasks such as clustering. Among its variants, the denoising autoencoders (DAEs) take partially corrupted inputs and are trained to recover original undistorted inputs~\citep{vincent2010stacked}. Variational autoencoders (VAEs) model the latent space with probabilistic models. As these probabilities are complex and usually intractable, they adopt a variational inference technique to approximate these probabilistic models~\citep{kingma2013auto}. 

\textit{Suitable biomedical data:} unlabeled data. 

\xhdr{Generative Models} In contrast to making a prediction, generative models aim to learn a sufficient statistical distribution that characterizes the underlying datasets (e.g., a set of DNA sequences for a disease) and its generation process~\citep{wittrock1974learning}. Based on the learned distribution, various kinds of downstream tasks can be supported. For example, from this distribution, one can intelligently generate optimized data points. These optimized samples can be novel images, compounds, or RNA sequences. One popular model is called generative adversarial networks (GAN)~\cite{goodfellow2014generative}. It consists of two sub-models: a {\it generator} that captures the data distribution of a training dataset in a latent representation and a {\it discriminator} that determines whether a sample is real or generated. These two sub-models are trained iteratively such that the resulting generator can produce realistic samples that potentially fool the discriminator. 

\textit{Suitable biomedical data:} data where new variants can have more desirable properties (e.g., molecule generation for drug discovery)~\citep{fu2020core,jin2018junction}. Depending on the data modality, different encoders can be chosen for the generative models.

\begin{table}[t]
    \centering
    \caption{High quality machine learning datasets references and pointers for genomics therapeutics tasks.}
    \adjustbox{max width=\textwidth}{
    \begin{tabular}{l|l|l|l}
    \toprule
    \textbf{Pipeline} & \textbf{Task} & \textbf{Data Reference}  & \textbf{Data Link} \\ \midrule 
    \multirow{10}{*}{Target Discovery (Sec.\ref{sec:target})} & DNA/RNA-protein binding & \cite{zeng2016convolutional} & \url{http://cnn.csail.mit.edu/} \\
    & Methylation state & \cite{levy2019pymethylprocess} & \url{https://bit.ly/3rVWgR9}\\
    & RNA splicing & \cite{harrow2012gencode} & \url{https://www.gencodegenes.org/}\\
    & Spatial gene expression & \cite{weinstein2013cancer} & \url{https://bit.ly/3fOLgTi}\\
    & Cell composition analysis & \cite{cobos2020benchmarking}& \url{https://go.nature.com/3mxCZEv}\\
    & Gene network construction & \cite{shrivastava2020grnular} & \url{https://bit.ly/3mBMB1f} \\
    & Variant calling & \cite{chen2019systematic} & \url{https://bit.ly/39RJcG6} \\
    & Variant prioritization  & \cite{landrum2014clinvar} & \url{https://www.ncbi.nlm.nih.gov/clinvar/} \\
    & Gene-disease association & \cite{pinero2016disgenet} & \url{https://www.disgenet.org/}\\
    & Pathway analysis & \cite{fabregat2018reactome}  & \url{https://reactome.org/}\\ \midrule 
     
     \multirow{5}{*}{Therapeutics Discovery (Sec.\ref{sec:discovery})} &  Drug response  & \cite{yang2012genomics}& \url{https://www.cancerrxgene.org/}\\
     & Drug combination & \cite{liu2020drugcombdb} & \url{http://drugcombdb.denglab.org/}\\
     & CRISPR on-target& \cite{leenay2019large} & \url{https://bit.ly/3rXlKxi} \\
     & CRISPR off-target& \cite{stortz2021crisprsql} & \url{http://www.crisprsql.com/}\\
     & Virus vector design& \cite{bryant2021deep}& \url{https://bit.ly/31RRKIP} \\ \midrule

     \multirow{4}{*}{Clinical Study (Sec.\ref{sec:clinical})} & Cross-species translation & \cite{poussin2014species}& \url{https://bit.ly/3mykFLC} \\
     & Patient stratification & \cite{curtis2012genomic} & \url{https://bit.ly/3cWTW8d} \\
     & Patient-trial matching & \cite{zhang2020deepenroll} & \url{https://bit.ly/3msp0A0}\\
     & Mendelian randomization & \cite{hemani2017automating} & \url{https://www.mrbase.org/}\\
\midrule 
     
     \multirow{2}{*}{Post-Market Study (Sec.\ref{sec:post-market})} & Clinical texts mining & Proprietary & N/A\\
     & Biomedical literature mining & \cite{pyysalo2007bioinfer} & \url{https://bit.ly/3cUtpYZ}\\ \midrule 
    \end{tabular}
    }
    
    \label{tab:database}
\end{table}

\section{Machine Learning for Genomics in Target Discovery} \label{sec:target}

A therapeutic target is a molecule (e.g., a protein) that plays a role in the disease biological process. The molecule could be targeted by a drug to produce a therapeutic effect such as inhibition, thereby blocking the disease process. Much of target discovery relies on fundamental biological research in depicting a full picture of human biology, and based on this knowledge, we identify target biomarkers. In this section, we review ML for genomics tasks in target discovery. In Section~\ref{sec:human_bio}, we review six tasks that use ML to facilitate understanding of human biology, and in Section~\ref{sec:biomarker}, we describe four tasks in using ML to help identify druggable biomarkers more accurately and more quickly.

\subsection{Facilitating Understanding of Human Biology} \label{sec:human_bio}

\begin{figure}[t]
    \centering
    \includegraphics[width = 0.8\textwidth]{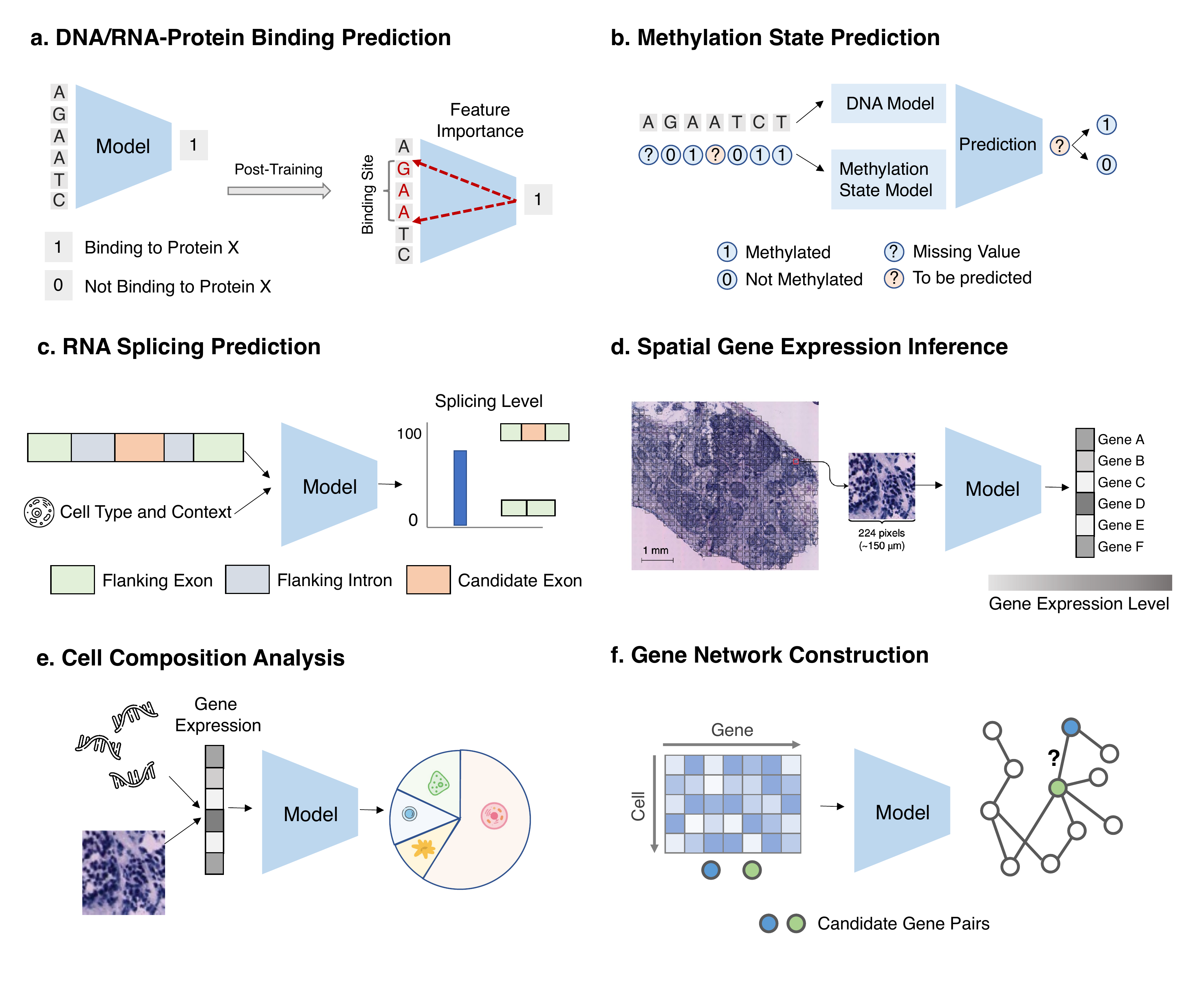}
    \caption{\textbf{Task illustrations for the theme "facilitating understanding of human biology"}. \textbf{a.} A model predicts if a DNA/RNA sequence can bind to a protein. After training, one can identify binding sites based on feature importance (Section~\ref{sec: dna-protein}). \textbf{b.} A model predicts missing DNA methylation state based on its neighboring states and DNA sequence (Section~\ref{sec:methy}). \textbf{c.} A model predicts the splicing level given the RNA sequence and the context (Section~\ref{sec:splice}). \textbf{d.} A model predicts spatial transcriptomics from tissue image (Section~\ref{sec:spatial}). \textbf{e.} A model predicts the cell type compositions from the gene expression (Section~\ref{sec:composition}). \textbf{f.} A model constructs a gene regulatory network from gene expressions (Section~\ref{sec:network_construction}).
    Credits: Figure c is adapted from, \cite{xiong2015human} and the spatial transcriptomics image in Figure d is from \cite{he2020integrating}.}
    \label{fig:human_biology}
\end{figure}

Oftentimes, the first step for developing any therapeutic agent is to generate a disease hypothesis and understand the disease mechanisms. It requires some understanding of basic human biology since diseases are complicated and driven by many factors. Machine learning applied to genomics can facilitate basic biomedical research and help understand disease mechanisms. A wide range of relevant tasks have been tackled by machine learning, from predicting splicing patterns~\citep{jha2017integrative,xiong2015human}, DNA methylation status~\citep{angermueller2017deepcpg}, to decoding the regulatory roles of genes~\citep{liu2016pedla,deepsea}. The majority of previous reviews have focused on this theme only. While there are numerous tasks under this category, we will describe just six important and popular tasks here. 

\subsubsection{DNA-protein and RNA-protein binding prediction} \label{sec: dna-protein}

DNA-binding proteins bind to specific DNA strands (binding sites/motifs) to influence the transcription rate to RNA, chromatin accessibility, and so on. These motifs regulate gene expression and, if mutated, can potentially contribute to diseases. Similarly, RNA-binding proteins bind to RNA strands to influence RNA processing, such as splicing and folding. Thus, it is important to identify the DNA and RNA motifs for these binding proteins. 

Traditional approaches are based on position weight matrices (PWMs), but they require existing knowledge about the motif length and typically ignore interactions among the binding site loci. Machine learning models trained directly on sequences to predict binding scores circumvent these challenges. \citep{alipanahi2015predicting} uses a convolutional neural network to train large-scale DNA/RNA sequences with varying lengths to predict the binding scores. The use of CNN is a great match for this task because the CNN’s filters work according to a similar mechanism as PWMs, which means that we can visualize binding site motifs through CNN filter weights. While motifs are useful, they have lower predictive power than evolutionary features~\citep{kircher2014general} for identifying chromatin proteins/histone marks binding. \citep{deepsea} shows that integrating another CNN model on additional information from the epigenomics profile could better predict these marks. Extending CNN-based models, a large body of works has been proposed to predict DNA-, RNA-protein binding~\citep{kelley2016basset,zhang2018high,zeng2016convolutional,cao2019simple}. 

\textit{Machine learning formulation: } Given a set of DNA/RNA sequences predict their binding scores. After training, use feature importance attribution methods to identify the motifs. Task illustration is in Figure~\ref{fig:human_biology}a.

\subsubsection{Methylation state prediction} \label{sec:methy}

DNA methylation adds methyl groups to individual A or C bases in the DNA to modify gene activity without changing the sequence. It has been shown to be a commonly used mediator for biological processes such as cancer progression and cell differentiation~\citep{robertson2005dna}. Thus, it is important to know the methylation status for DNA sequences in various cells. However, since the single-cell methylation technique has low coverage, most of the methylation status at specific DNA positions is missing, so it requires accurate imputation. 

Classical methods can only predict population-level status instead of cell-level as cell-level prediction require annotations that are unavailable~\citep{zhang2015predicting,whitaker2015predicting}. Machine learning models can tackle this problem. Given a set of cells with their available sequenced methylation status for each DNA position and the DNA sequence, \citep{angermueller2017deepcpg} accurately infers the unmeasured methylation statuses in a single-cell level. More specifically, the imputation of DNA methylation positions uses a bidirectional recurrent neural network on a sequence of cells' neighboring available methylation states and a CNN on the DNA sequence. The combined embedding takes into account information between DNA and methylation status across cells and within cells. Alternative architecture choice has also been proposed, such as using Bayesian clustering~\citep{kapourani2019melissa}, or a variational auto-encoder~\citep{levy2020methylnet}. Notably, it can also be extended to RNA methylation state prediction. \citep{zou2019gene2vec} applies CNN on the neighboring methylation status and the word2vec model on RNA subsequence for RNA methylation status prediction. 

\textit{Machine learning formulation: } For a DNA/RNA position with missing methylation status, given its available neighboring methylation states and the DNA/RNA sequence, predict the methylation status on the position of interest. Task illustration in Figure~\ref{fig:human_biology}b.

\subsubsection{RNA splicing prediction}  \label{sec:splice}

RNA splicing is a mechanism to assemble the coding regions and remove the non-coding ones to be translated into proteins. A single gene can have various functions by splicing the same gene in different ways given different conditions. \cite{lopez2005splicing} estimates that as many as 60\% of pathogenic variants responsible for genetic diseases may influence splicing.  \cite{gelfman2017annotating} used ML to derive a score, TraP, which identifies around 2\% of synonymous variants and 0.6\% of intronic variants as likely pathogenic due to splicing defects. Thus, it is important to be able to identify the genetic variants that cause splicing.

\cite{xiong2015human} models this problem as predicting the splicing level of an exon, measured by the transcript counts of this exon, given its neighboring RNA sequence and the cell type information. It uses Bayesian neural network ensembles on top of curated RNA features and has demonstrated its accuracy by identifying known mutations and discovering new ones. Notably, this model is trained on large-scale data across diverse disease areas and tissue types. Thus, the resulting model can predict the effect of a new unseen mutation contained within hundreds of nucleotides on splicing of an intron without experimental data. In addition, to predict the splicing level given a triplet of exons in various conditions, recent models have been developed to annotate the nucleotide branchpoint of RNA splicing. \cite{paggi2018sequence} feeds an RNA sequence into a recurrent neural network, where it predicts the likelihood of being a branchpoint for each nucleotide. \cite{jagadeesh2019s} further improve the performance by integrating features from the splicing literature and generate a highly accurate splicing-pathogenicity score. 

\textit{Machine learning formulation: } Given an RNA sequence and its cell type, if available, for each nucleotide, predicts the probability of being a spliced breakpoint and the splicing level. Task illustration is in Figure~\ref{fig:human_biology}c.



\subsubsection{Spatial gene expression inference} \label{sec:spatial}

Gene expression varies across the spatial organization of tissue. This heterogeneity contains important insights about the biological effects. Regular sequencing, whether of single-cells or bulk tissue, does not capture this information. Recent advances in spatial transcriptomics (ST) characterize gene expression profiles in their spatial tissue context~\citep{staahl2016visualization}. However, there are still challenges to integrating the sequencing output with the tissue context provided by histopathology images to better visualize and understand patterns of gene expression within a tissue section. 

Machine learning models that directly predict gene expression from the histopathology image can thus be a useful tool. \cite{he2020integrating} develops a deep CNN that predicts gene expression from histopathology of patients with breast cancer at a resolution of 100 $\mu$m. They also show the model can generalize to other breast cancer datasets without re-training. Building upon the inferred spatial gene expression levels, many downstream tasks are enabled. For example, \cite{levy2020spatial} constructs a pipeline that characterizes tumor heterogeneity on top of the CNN gene expression inference step. 

\textit{Machine learning formulation: } Given the histopathology image of the tissue, predict the gene expression for every gene at each spatial transcriptomics spot.  Task illustration is in Figure~\ref{fig:human_biology}d.

\subsubsection{Cell composition analysis}  \label{sec:composition}

Different cell types can drive change in gene expressions that are unrelated to the interventions. Analyzing the average gene expression for a batch of mixed cells with distinct cell types could lead to bias and false results~\citep{egeblad2010tumors}. Thus, it is important to deconvolve the gene expressions of the cell-type composition from the real signals for tissue-based RNA-seq data. 

ML models can help estimate the cell type proportions and the gene expression. The rationale is to obtain parameters of gene expression (a signature matrix) that characterize each cell type through single-cell profiles. The signature matrix should contain gene expressions that are stably expressed across conditions. These parameters are then integrated into the RNA-seq data to infer cell composition for a set of query gene expression profiles. Various methods, including linear regression~\citep{avila2018computational} and support vector machines~\citep{newman2015robust} are used to predict a cell composition vector when combined with the signature matrix to approximate the gene expression. In these cases, the signature matrix is predefined, which may not be optimal. \cite{menden2020deep} applies DNNs to predict cell composition profile directly from the gene expression, where the hidden neurons can be considered as the learned signature matrix. Cell deconvolution is also crucial for spatial transcriptomes where each spot could contain 2 to 20 cells from a mixture of dozens of possible cell types. \cite{andersson2020single} models various cell type-specific parameters using a customized probabilistic model. \cite{su2020dstg} uses graph convolutional network to leverage information from similar spots in the spatial transcriptomics. However, this problem is constrained by the limited availability of gold standard cell composition annotations. 

\textit{Machine learning formulation: } Given the gene expressions of a set of cells (in bulk RNA-seq or a spot in spatial transcriptomics), infer proportion estimates of each cell type for this set. Task illustration in Figure~\ref{fig:human_biology}e.

\subsubsection{Gene network construction} \label{sec:network_construction}

The expression levels of a gene are regulated via transcription factors (TF) produced by other genes. Aggregating these TF-gene relations results in the gene regulatory network. Accurate characterization of this network is crucial because it describes how a cell functions. However, it is difficult to quantify gene networks on a large-scale through experiments alone. 

Computational approaches have been proposed to construct gene networks from gene-expression data. The majority of them learn a mapping from expressions of a gene to TF. If the mapping is successful, then it is likely that this TF affects this gene. Various mapping methods have been proposed, such as linear regression~\citep{haury2012tigress}, random forest~\citep{huynh2010inferring}, and gradient boosting~\citep{moerman2019grnboost2}. \cite{shrivastava2020grnular} proposes a deep neural network version of the mapping through a specialized unrolled algorithm to control the sparsity of the learned network. They also leverage supervision obtained through synthetic data simulators to improve robustness further. Despite the promises, this problem remains unsolved due to the sparsity, heterogeneity, and noise of the gene expression data, particularly data from single cell RNA sequencing. 

\textit{Machine learning formulation: } Given a set of gene expression profiles of a gene set, identify the gene regulatory network by predicting all pairs of interacting genes. Task illustration is in Figure~\ref{fig:human_biology}f.

\subsection{Identifying Druggable Biomarkers} \label{sec:biomarker}

\begin{figure}[t]
    \centering
    \includegraphics[width=0.9\textwidth]{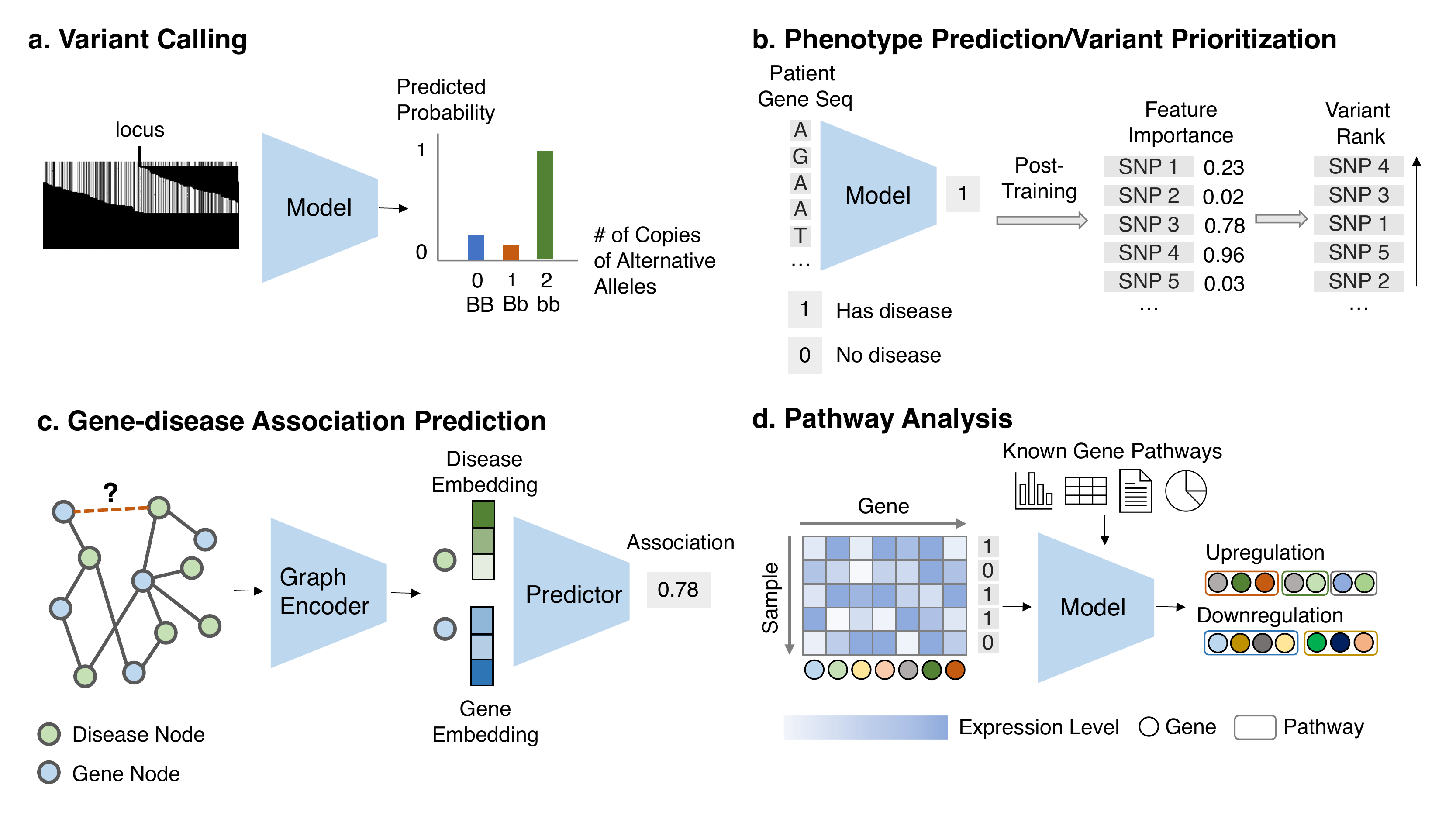}
    \caption{\textbf{Task illustrations for the theme "identifying druggable biomarkers".} \textbf{a.} A model predicts the zygosity given a read pileup image (Section~\ref{sec:calling}). \textbf{b.} A model predicts whether this patient has the disease given the genomic sequence. After training, feature importance attribution methods are used to assign importance for each variant, which is then ranked and prioritized (Section~\ref{sec:variant_prior}). \textbf{c.} A graph encoder obtains embeddings for each disease and gene node, and they are fed into a predictor to predict their association (Section~\ref{sec:gda}). \textbf{d.} A model identifies a set of gene pathways from the gene expression profiles and the known gene pathways (Section~\ref{sec:pathway}).}
    \label{fig:biomarker}
\end{figure}

Diseases are driven by complicated biological processes where each step may be associated with a biomarker. By identifying these biomarkers, we can design therapeutics to break the disease pathway and cure the disease. Machine learning can help identify these biomarkers by mining through large-scale biomedical data to predict genotype-phenotype associations accurately. Probing the trained models can uncover potential biomarkers and identify patterns related to the disease mechanisms. Next, we will present several important tasks related to biomarker identification.

\subsubsection{Variant calling}  \label{sec:calling}
Variant calling is the very first step before relating genotypes to diseases. It is used to specify what genetic variants are present in each individual’s genome from sequencing. The majority of the variants are biallelic, meaning that each locus has only one possible alternative form of nucleotide compared to the reference, while a small fraction are also multiallelic, meaning that each locus can have more than one alternate form. As each locus has two copies, one from mother and another from father, the variant is measured by the total set of nucleotides (e.g., for biallelic variant, suppose B is the reference nucleotide, and b is the alternative; three genotypes are possible: homozygous (BB), heterozygous (Bb) and homozygous alternate(bb)). Raw sequencing outputs are usually billions of short reads, and these reads are aligned to a reference genome. In other words, for each locus, we have a set of short reads that contain this locus. Since sequencing techniques have errors, the challenge is to predict the variant status of this locus accurately from the set of reads. Manual processing of such a large number of reads to identify each variant is infeasible. Thus, efficient computational approaches are needed for this task. 

A statistical framework called the Genome Analysis Toolkit (GATK)~\citep{depristo2011framework}, combines logistic regression, hidden Markov models, and Gaussian mixture models, and is commonly used for variant calling. Deep learning methods have shown improved performance. For example, while previous works operate on sequencing statistics, DeepVariant~\citep{poplin2018universal} treats the sequencing alignments as an image and applies CNNs. It has been shown to have superior performance to previous modeling efforts and also works for multiallelic variant calling. In addition to predicting zygosity, \cite{luo2019multi} use multi-task CNNs to predict the variant type, alternative allele, and indel length. Many other deep learning based methods are proposed to tackle more specific challenges, such as long sequencing length using LSTMs~\citep{luo2020exploring}. Benchmarking efforts have also been conducted~\citep{zook2019open}. Note that despite most methods achieving greater than 99\% accuracy, thousands of variants are still being called incorrectly since the genome sequence is extremely long. Also, variability persists across different sequencing technologies. Another challenge is the phasing problem, which is to estimate whether the two mutations in a gene are on the same chromosome (haplotypes) or opposite ones~\citep{delaneau2013improved}. Thus, there is still room for further improvement. 

\textit{Machine learning formulation: } Given the aligned sequencing data ((1) read pileup image, which is a matrix of dimension $M$ and $N$, with $M$ the number of reads and $N$ the length of reads; or (2) the raw reads, which are a set of sequences strings) for each locus, classify the multi-class variant status.  Task illustration is in Figure~\ref{fig:biomarker}a.

\subsubsection{Variant pathogenicity prioritization/phenotype prediction} \label{sec:variant_prior}

There are an extensive number of genomic variants in the human genome, at last one million per person. While many influence complex traits and are relatively harmless, some are associated with diseases. Complex diseases are associated with multiple variants in both coding and non-coding regions of the genome. Thus, prioritization of pathogenic variants from the entire variant set can potentially lead to disease targets. 

There are mainly two computational approaches. The first one is to predict the pathogenicity given a set of features for a single variant. These features are usually curated from biochemical knowledge, such as amino acid identities.  \cite{kircher2014general} build on these features using a linear support vector machine and \cite{quang2015dann} use deep neural networks to classify if a variant is pathogenic. DNN shows improved performance on classification metrics. After training, the model can generate a ranked list of variants based on their predicted pathogenicity likelihood where the top ones are prioritized. Note that this line of work considers each variant as an input data point and assumes some known knowledge of the pathogenicity of the variants, which is not the case in many scenarios, especially for new diseases. 

Another line of work is to use each genome profile as a data point and use a computational model to predict disease risks from this profile. If the model is accurate, one can obtain variants contributing to the prediction of the disease phenotype. Predicting directly from the whole-genome sequence is challenging because of two reasons. First, as the whole-genome is high-dimensional while the cohort size for each disease is relatively limited, this presents the "curse of dimensionality" challenge in machine learning. Secondly, most SNPs in the input genome are irrelevant to the disease, presenting difficulty in correctly identifying these signals from the noise. \cite{kooperberg2010risk} uses a sparse regression model to predict the risk of Crohn's disease for patients using genomics data in the coding region. \cite{pare2017machine} uses gradient boosted regression to approximate polygenic risk score for complex traits such as diabetes, height, and BMI. \cite{isgut2021highly} uses logistic regression on polygenic risk scores to improve myocardial infarction risk prediction. \cite{zhou2018deep} applies DNNs on the epigenomic features of both the coding and non-coding regions to predict gene expression for more than 200 tissue and cell types and later identify disease-causing SNPs. Built upon DeepSEA~\citep{deepsea}, \cite{zhou2019whole} apply CNN on epigenomic profiles, which are modifications of the DNA sequence such as DNA methylation or chromatin accessibility, to predict autism and identify experimentally validated non-coding variant mutations. 

\textit{Machine learning formulation: } Given features about a variant, predict its corresponding disease risk and then rank all variants based on the disease risk. Alternatively, given the DNA sequence or other related genomics features, predict the likelihood of disease risk for this sequence and retrieve the variant in the sequence that contributes highly to the risk prediction. Task illustration is in Figure~\ref{fig:biomarker}b.

\subsubsection{Rare disease detection} \label{sec:rare}

In the US, a rare disease is defined as one that affects fewer than 200,000 people, with other countries similarly defining a rare disease based on low prevalence. There are around 7,000 rare diseases, and they collectively affect 350 million people worldwide~\citep{vickers2013challenges}. Due to limited financial incentives, unknown disease mechanisms and potential difficulties in recruiting sufficient patients for clinical trials, more than 90\% of rare diseases lack effective treatments. Also, initial misdiagnosis is common. On average, it takes more than seven years and eight physicians for a patient to be correctly diagnosed. Importantly, it is likely that targets identified for rare diseases may also be useful for therapeutic intervention of similar more common diseases.

ML models are good at identifying patterns from complex patient data. Rare disease detection can be formulated as a classification task, similar to phenotype prediction. It aims to identify if the patient has a rare disease from the patient's genomic sequence and information such as EHR. If sufficient data from patients with a rare disease and suitable controls exist, many ML models can be applied to detect rare diseases. For example, based on the motivation that many rare diseases have missing heritability, which could be harbored in regulatory regions, \cite{yin2019using} propose a two-step CNN approach where one CNN first predicts the promoter regions that are likely associated with Amyotrophic Lateral Sclerosis. Another CNN detects if the patient has the rare disease based on genotypes in the selected genomic regions. 

However, rare diseases pose special challenges to ML compared to classical phenotype prediction because these diseases have an extremely low prevalence in the data while the majority of data points belong to the control set. This data imbalance makes it difficult for ML models to pick up signals and hence prevent them from making an accurate prediction. Thus, special model designs are required. \cite{cui2020conan} uses a generative adversarial network (GAN) model to generate synthetic but realistic rare disease patient embeddings to alleviate the class imbalance problem and show significant performance increase in rare disease detection. \cite{taroni2019multiplier} use a transfer learning framework to adapt from large-scale genomic data with a diverse set of diseases to a smaller set of rare disease genomic data. Specifically, they leverage biological principle by constructing latent variables shared across a wide range of diseases. These variables correspond to genetic pathways. As these variables are the fundamental biology units, they can be naturally adopted even for smaller datasets such as rare disease cohorts.

\textit{Machine learning formulation: } Given the gene expression data and other auxiliary data of a patient predict whether this patient has a rare disease. Also, identify genetic variants for this rare disease. Task illustration is in Figure~\ref{fig:biomarker}b, which is the same as phenotype prediction.

\subsubsection{Gene-disease association prediction} \label{sec:gda}

Although numerous genes are now mapped to diseases, human knowledge about gene-disease association mapping is vastly incomplete. At the same time, we know many genes are similar to each other, as is also the case for diseases. We can impute unknown associations from known ones by many similarity rules that govern the gene-disease networks to leverage these similarities. One notable rule is the "guilt by association" principle~\citep{wolfe2005systematic}. For example, disease $X$ and gene $a$ are more likely to be associated if we know gene $b$ associated with disease $X$ has a similar functional role as gene $a$. In contrast to variant prioritization focusing on prediction of one specific disease, gene-disease association predictions aim to predict any disease-gene pairs. 

Many graph-theoretic approaches such as diffusion~\citep{kohler2008walking} have been applied to gene-disease association prediction. However, they require strong assumptions about the data. Learnable methods have also been heavily investigated. This problem is also being formulated as a recommendation system problem where it recommends items(genes) to users(diseases). \cite{huang2020skipgnn} use a molecular network-motivated graph neural network and formulate association prediction as a link prediction problem. Studies have shown that integrating similarity across multiple data types can help gene-disease prediction~\citep{tranchevent2016candidate}. Thus, a multi-modal data fusion scheme is also ideal. Notably, \cite{luo2019enhancing} fuse information from protein-protein interaction and gene ontology through a multimodal deep belief network. As some diseases are not well annotated compared to others, predicting molecularly uncharacterized (no known biological function or genes) diseases such as rare diseases is also important. \cite{caceres2019disease} use phenotype data to transfer knowledge from other phenotypically similar diseases using a network diffusion method, where the phenotypical similarity is defined by the distance on the disease ontology trees.

\textit{Machine learning formulation: } Given the known gene-disease association network and auxiliary information, predict the association likelihood for every unknown gene-disease pair. Task illustration is in Figure~\ref{fig:biomarker}c.

\subsubsection{Pathway analysis and prediction} \label{sec:pathway}

Many diseases are driven by a set of genes forming disease pathways. Pathway analysis identifies these gene sets through transcriptomics data and leads toward a more complete understanding of disease mechanisms. Many statistical approaches have been proposed. For example, Gene Set Enrichment Analysis~\citep{subramanian2005gene} leverages existing known pathways and calculates statistics on omics data to see if any pathway is activated. However, it treats each pathway as a set, while no relation among the genes is provided. Other topology-based pathway analyses~\citep{tarca2009novel} that take into account the gene relational graph structure are also proposed. Many pathway analyses suffer from noise and provide unstable pathway activation and inhibition patterns across samples and experiments. 
\cite{ozerov2016silico} introduces a clustered gene importance factor to reduce noise and improve robustness. Although current pathway analysis still heavily relies on network-based methods~\citep{reyna2020pathway}, an emerging trend used to understand potential disease mechanisms is to probe into explainable machine learning models that predict genotype-to-disease association. Many efforts have been made to simulate cell signaling pathways and corresponding hierarchical biological processes \textit{in silico}. \cite{karr2012whole} devises the first whole-cell approach to predict cell growth from genotype using a set of differential equations. Recently, a machine learning model called visible neural network~\citep{ma2018using} simulates the hierarchical biological processes (gene ontology) in a eukaryotic cell as a feedforward neural network where each neuron corresponds to a biologic subsystem. This model is trained end-to-end from genotype to cell fitness phenotype with good accuracy. A post-hoc interpretability method that assigns scores for each subsystem generates a likely mechanism for the fitness of a cell after training. This method has been extended recently to train on genomics data related to prostate cancer phenotype, in order to generate disease pathways~\citep{elmarakeby2020biologically}. 

\textit{Machine learning formulation: } Given the gene expression data for a phenotype and known gene relations, identify a set of genes corresponding to disease pathways. Task illustration is in Figure~\ref{fig:biomarker}d.

\section{Machine Learning for Genomics in Therapeutics Discovery} \label{sec:discovery}

After a drug target is identified, a campaign to design potent therapeutic agents to modulate the target and block the disease pathway is initiated. These therapeutics can be a small molecule, an antibody, gene therapy, and so on. The discovery consists of numerous phases and subtasks to ensure the efficacy and safety of the therapeutics. Genomics data also play a role in this process. In this section, we review ML for genomics in therapeutics discovery under two main themes. Section~\ref{sec:personalized} investigates the relation of small-molecule drug efficacy given different cellular genomic contexts. Section~\ref{sec:gene_therapy} reviews how ML can enable the design of various gene therapies.

\subsection{Improving Context-specific Drug Response}\label{sec:personalized}

\begin{figure}[t]
    \centering
    \includegraphics[width=0.9\textwidth]{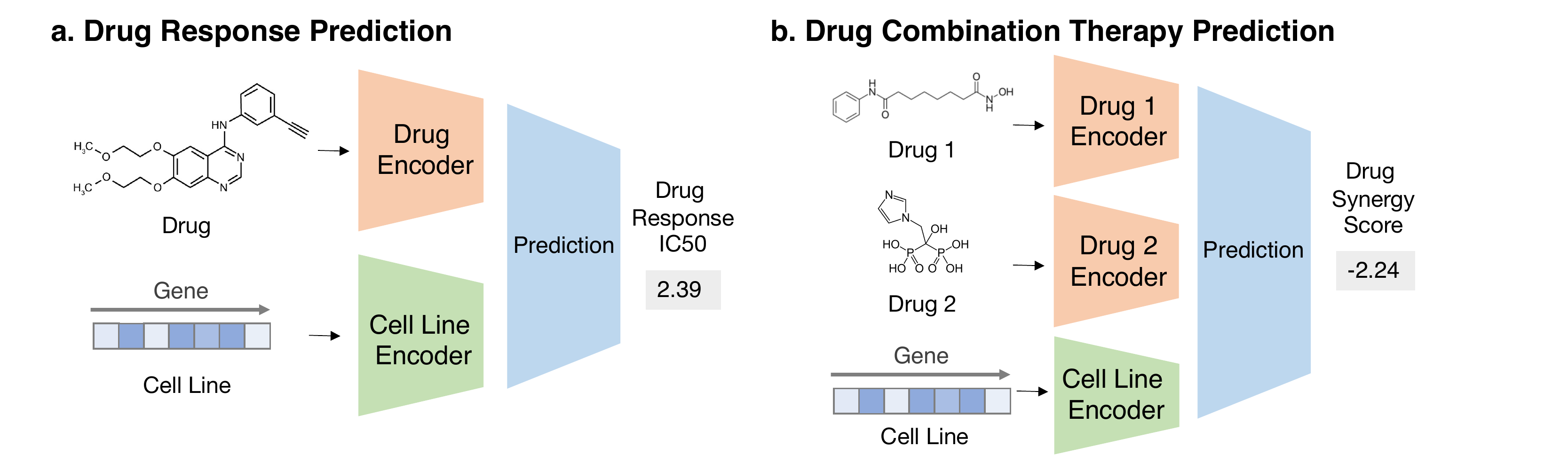}
    \caption{\textbf{Task illustrations for the theme "improving context-specific drug response"}. \textbf{a.} A drug encoder and a cell line encoder produce embeddings for drug and cell line, respectively, which are then fed into a predictor to estimate drug response (Section~\ref{sec:drug_response}). \textbf{b.} Drug encoders first map two drugs into embedding, and a cell line encoder maps a cell line into embeddings. Then, three embeddings are fed into a predictor for drug synergy scores (Section~\ref{sec:drug_combo}). }
    \label{fig:personalized}
\end{figure}

Personalized medicine aims at developing the treatment strategy based on a patient's genetic profile. This contrasts with the traditional "one-size-fits-all" approach, which assigns the same treatments to patients with the same diseases. Personalized approaches have been one of the most sought-after endeavors in the field due to their numerous advantages such as improving outcomes and reducing side effects~\citep{hamburg2010path}, especially in oncology, where several biomarkers could lead to drastically different treatment plans~\citep{chin2011cancer}. Despite the promise to understand the relations among treatments, diseases, high-dimensional genomics profiles, and the various outcomes, large-scale experiments in combinatorial complexity are required to investigate these relationships~\citep{menden2019community}. Machine learning provides valuable tools to facilitate this process. 

\subsubsection{Drug response prediction} \label{sec:drug_response}

It is known that the same small-molecule drug could have various response levels given different genomic profiles. For example, an anti-cancer drug has a different response to different tumors. Thus, it is crucial to generate an accurate response profile given drug-genomics profile pairs. However, to experimentally test each combination of available drugs and cell-line genomics profiles is prohibitively expensive. 

A machine learning model can be used to predict a drug's response in a diverse set of cell lines \textit{in silico}. An accurate machine learning model can greatly narrow down the drug screening space and reduce the burden on experimental costs and resources. Various models have been proposed to improve the accuracy, such as matrix factorization~\citep{ammad2016drug}, VAEs~\citep{rampavsek2019dr}, ensemble learning~\citep{tan2019drug}, similarity network model~\citep{zhang2015predicting2}, and feature selection~\citep{ali2019machine}. While promising, one challenge is that the current public database has a limited number of drugs and genomics profiles tested, especially for some tissues or drug classes. It is unclear if the model can generalize to new contexts such as novel cell types and structurally diverse drugs with limited samples. To tackle this challenge, \cite{ma2021few} apply model-agnostic meta-learning to learn from screening data of a set of tissues to generalize to new contexts such as new tissue types and preclinical studies in mice~\citep{finn2017model}. In addition to accurate prediction, it is also important to allow an understanding of drug response mechanisms. \citep{kuenzi2020predicting} Applying visible neural networks in the drug response prediction context, \cite{ma2018using} generates potential mechanisms and validated them through experiments using CRISPR, in-vitro screening, and patient-derived tissue cultures. 

\textit{Machine learning formulation: } Given a pair of drug compound molecular structure and gene expression profile of the cell line, predict the drug response in this context. Task illustration is in Figure~\ref{fig:personalized}a.

\subsubsection{Drug combination therapy prediction}  \label{sec:drug_combo}

Drug combination therapy, also called cocktails, can expand the use of existing drugs, improve outcomes, and reduce side effects. For example, drug cocktails can modulate multiple targets to provide a novel mechanism of action in cancer treatments. Also, by reducing dosages for each drug, it may be possible to reduce adverse effects. However, screening the entire space of possible drug combinations and various cell lines is not feasible experimentally. 

Machine learning that can predict synergistic responses given the drug pair and the genomic profile for a cell line can prove valuable. Classical machine learning methods such as naive Bayes~\citep{li2015large} and random forests~\citep{wildenhain2015prediction} have shown initial success on independent external data. Deep learning methods such as deep neural networks~\citep{preuer2018deepsynergy} and deep belief networks ~\citep{chen2018predict} have shown improved performance. Integration with multi-omics data on cell lines has also further improved the performance, such as miRNA expression and proteomic features~\citep{xia2018predicting}. Similar to drug response prediction, one important challenge is to transfer across tissue types and drug classes. \cite{kim2021anticancer} conducts transfer learning to adapt models trained on data-rich tissues such as brain and breast tissues to understudied tissues such as bone and prostate tissues. 

\textit{Machine learning formulation: } Given a combination of drug compound structures and a cell line's genomics profile, predict the combination response. Task illustration is in Figure~\ref{fig:personalized}b.




\subsection{Improving Efficacy and Delivery of Gene Therapy} \label{sec:gene_therapy}

\begin{figure}[t]
    \centering
    \includegraphics[width = 0.85\textwidth]{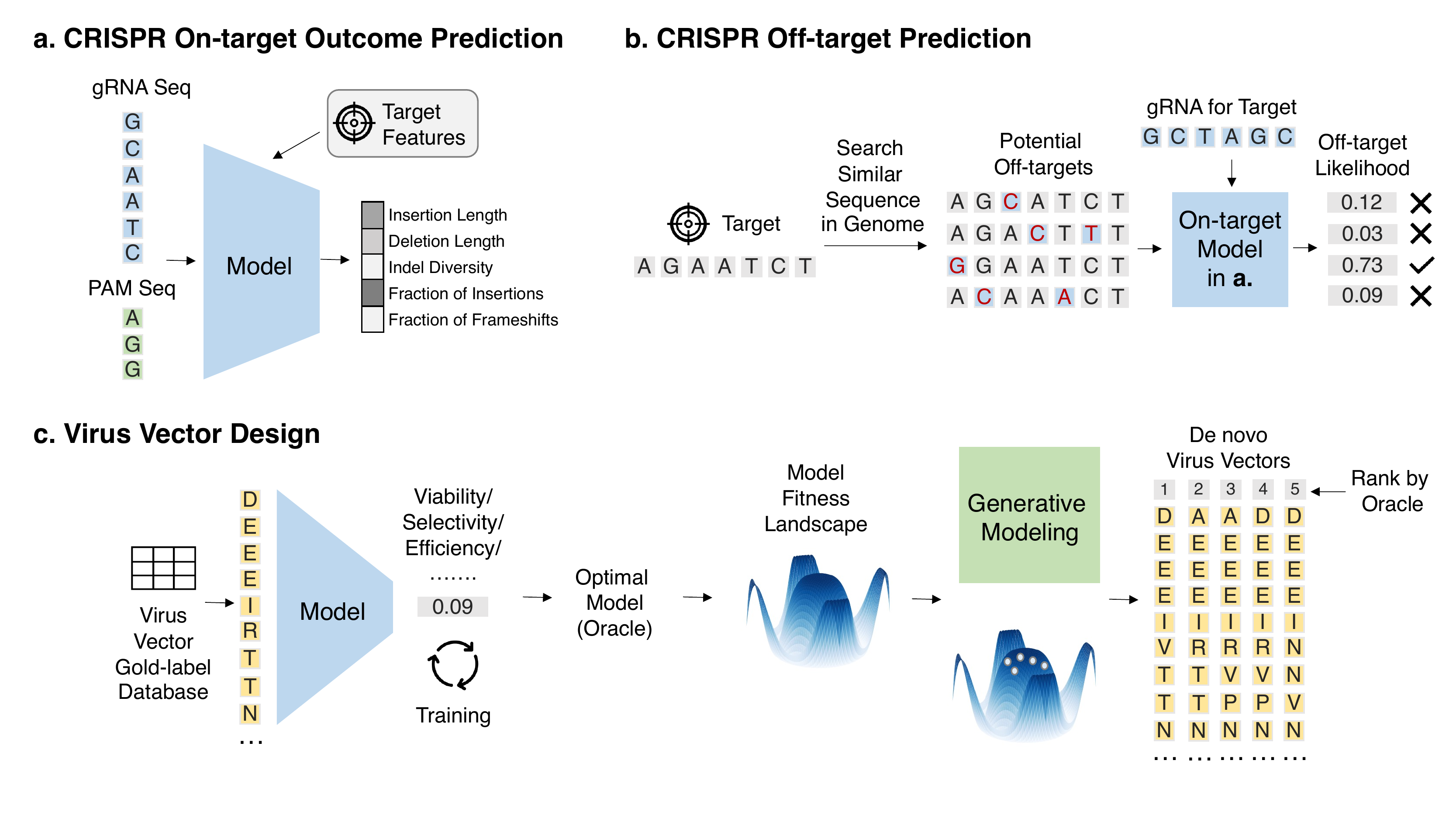}
    \caption{\textbf{Task illustrations for the theme "Improving Efficacy and Delivery of Gene Therapy".} \textbf{a.} A model predicts various gene editing outcomes given the gRNA sequence and the target DNA features (Section~\ref{sec:on_target}). \textbf{b.} First, a model search through similar sequences to the target DNA sequence in the candidate genome and generate a list of potential off-target DNA sequences. Next, an on-target model predicts if the gRNA sequence can affect these potential DNA sequences. The ones that have high on-target effects are considered potential off-targets (Section~\ref{sec:off_target}). \textbf{c.} An optimal model (oracle function) is first obtained by training on a gold-label database. Next, a generative model generates de novo virus vectors potent in the oracle fitness landscape (Section~\ref{sec:virus}). }
    \label{fig:gene}
\end{figure}

Gene therapy is an emerging therapeutics class, which delivers nucleic acid instruction into patient cells to prevent or cure disease. These instructions include (1) replacing disease-causing genes with healthy ones, (2) turning off genes that cause diseases, (3) inserting genes to produce disease-fighting proteins. Special vehicles called vectors are used to deliver these instructions (cargos) into the cells and induce sufficient therapeutic effects. Many choices exist, such as naked DNA, virus, and nanoparticles, and so on. Virus vectors have become popular due to their natural ability to directly enter cells and replicate their genetic material. Despite the promise, numerous challenges still exist in reaching the expected effect, such as the host immune response, viral vector toxicity, and off-target effects. In recent years, machine learning tools have been shown to help tackle many of these challenges. 

\subsubsection{CRISPR on-target outcome prediction} \label{sec:on_target}

CRISPR-Cas9 is a biotechnology that can edit genes in a precise location. It allows the correction of genetic defect to treat disease and provides a tool with which to alter the genome and to study gene function. CRISPR-Cas9 is a system with two important players. Cas9 protein is an enzyme that can cut through DNA, where the CRISPR sequence guides the cut location. The guide RNA sequence (gRNA) determines the specificity for the target DNA sequence in the CRISPR sequence. While existing CRISPR mostly make edits by small deletions, it is also of active research to do repairing, which after cutting, a DNA template is provided to fill in the missing part of the gene. In theory, CRISPR can correctly edit the target DNA sequence and even restore a normal copy, but in reality, the outcome varies significantly given different gRNAs~\citep{cong2013multiplex}. It has been shown that the outcome is decided by factors such as gRNA secondary structure and chromatin accessibility~\citep{jensen2017chromatin}. Some of the desirable outcomes include insertion/deletion length, indel diversity, the fraction of insertions/frameshifts. Thus, it is crucial to design a gRNA sequence such that the CRISPR-Cas system can achieve its effect on the designated target (also called on-target). 

Machine learning methods that can accurately predict the on-target outcome given the gRNA would facilitate the gRNA design process. Many classic machine learning methods have been investigated to predict various repair outcomes given gRNA sequence, such as linear models~\citep{labuhn2018refined,moreno2015crisprscan}, support vector machines~\citep{chari2015unraveling}, and random forests~\citep{wilson2018high}. However, they do not capture the high-order nonlinearity of gRNA features. Deep learning models that apply CNNs to automatically learn gRNA features show further improved performance~\citep{chuai2018deepcrispr,kim2018deep}. Numerous challenges still exist. For example, machine learning models are data-hungry. Limited data of CRISPR knockout experiments from the diverse cell and tissue types exist, affecting the model's generalizability. In particular, improving generalizability to novel target classes and generating prediction mechanisms are still an open question. 

\textit{Machine learning formulation: } With a fixed target, given the gRNA sequence and other auxiliary information such as target gene expression and epigenetic profile, predict its on-target repair outcome. Task illustration is in Figure~\ref{fig:gene}a.

\subsubsection{CRISPR off-target prediction} \label{sec:off_target}

As CRISPR can cut any region that matches the gRNA, it can potentially cut through similar off-target regions, leading to significant adverse effects. This is a major hurdle for CRISPR techniques for clinical implementations~\citep{zhang2015off}. Similar to on-target prediction, the off-target prediction is to predict if gRNA could cause off-target effects. In contrast to on-target, where we have a fixed given DNA region, off-target prediction requires identifying potential off-target regions from the entire genome. Thus, the first step is to search and narrow down a set of potential hits using alignment algorithms and distance measures~\citep{heigwer2014crisp,bae2014cas}. Next, given the set of targets and the gRNA, a model needs to score the putative target-gRNA pair.The model also needs to aggregate these scores since one gRNA usually has multiple putative off-targets. Various heuristics aggregation methods have been proposed and implemented~\citep{hsu2013dna,haeussler2016evaluation,cradick2014cosmid}. 

Machine learning methods improve performance further. \cite{listgarten2018prediction} uses a two-layer boosted regression tree where the first layer scores each gRNA-target pairs and the second layer aggregates the scores. \cite{lin2018off} apply CNN on a fused DNA-gRNA pair representation and achieve improved performance. Space for further improvement is large. For example, as data of richer contexts such as different cell, tissue, and organism types become available, more sophisticated models that can generalize well on all contexts could be possible.

\textit{Machine learning formulation: } Given the gRNA sequence and the off-target DNA sequence, predict its off-target effect. Task illustration is in Figure~\ref{fig:gene}b.

\subsubsection{Virus vector design} \label{sec:virus}

To deliver gene therapy instructions into cells and induce therapeutic effects, virus vectors are used as vehicles. The design of the virus vector is thus crucial. The recent development of Adeno-Associated Virus (AAV) capsid vectors leads to a surge in gene therapy due to its favorable tropism, immunogenicity, and manufacturability properties~\citep{daya2008gene}. However, there are stills unsolved challenges, mainly regarding the undesirable properties of natural AAV forms. For example, up to 50-70\% of humans are immune to the natural AAV vector, which means the human immune system would destroy it without delivering it to the targeted cells~\citep{chirmule1999immune}. This means that those patients are not able to receive gene therapies. Thus, designing functional variants of AAV capsids that can escape the immune system is crucial. Similarly, it would be ideal to design AAV variants that have higher efficiency and selectivity to the tissue target of interest. 

The standard method to generate new AAV variants is through "directed evolution" with limited diversity, most still similar to natural AAV. But this is very time- and resource-intensive, while the resulting yields are also low (<1\%).  Recently, \cite{bryant2021deep} developed a machine learning-based framework to generate AAV variants that can escape the immune system with a >50\% yield rate. They first train an ensemble neural network that aggregates DNN, CNN, and RNN using customized data collection to assign accurate viability scores given an AAV from diverse sources. Then, they sample iteratively on the predictor viability landscape to obtain a set of highly viable AAVs. Many opportunities remain open for machine-aided AAV design~\citep{kelsic2019challenges}. For example, this framework can be easily extended to other targets in addition to the immune system viability, such as tissue selectivity, if a high capacity machine learning property predictor can be constructed. 

\textit{Machine learning formulation: } Given a set of virus sequences and their labels for a property X, obtain an accurate predictor oracle and conduct various generation modeling to generate de novo virus variants with a high score in X and high diversity. Task illustration is in Figure~\ref{fig:gene}c.

\section{Machine Learning for Genomics in Clinical Studies} \label{sec:clinical}

After a therapeutic is shown to have efficacy in the wet lab, it is further evaluated in animals and then on humans in full-scale clinical trials. ML can facilitate this process using genomics data. We review the following three themes. Section~\ref{sec:translation} studies the long-standing problem of difficulty translating results from animals to humans and shows ML can enable better translation by better characterization of the molecular differences. Section~\ref{sec:cohort} reviews ML techniques to curate a better patient cohort that the therapeutic can be applied to, as it can greatly affect the clinical trial outcome. Section~\ref{sec:causal} surveys alternative ML techniques called causal inference to augment clinical trials in cases that traditional trials are not ethical or are difficult to conduct.

\subsection{Translating Preclinical Animal Models to Humans} \label{sec:translation}

\begin{figure}
    \centering
    \includegraphics[width=0.45\textwidth]{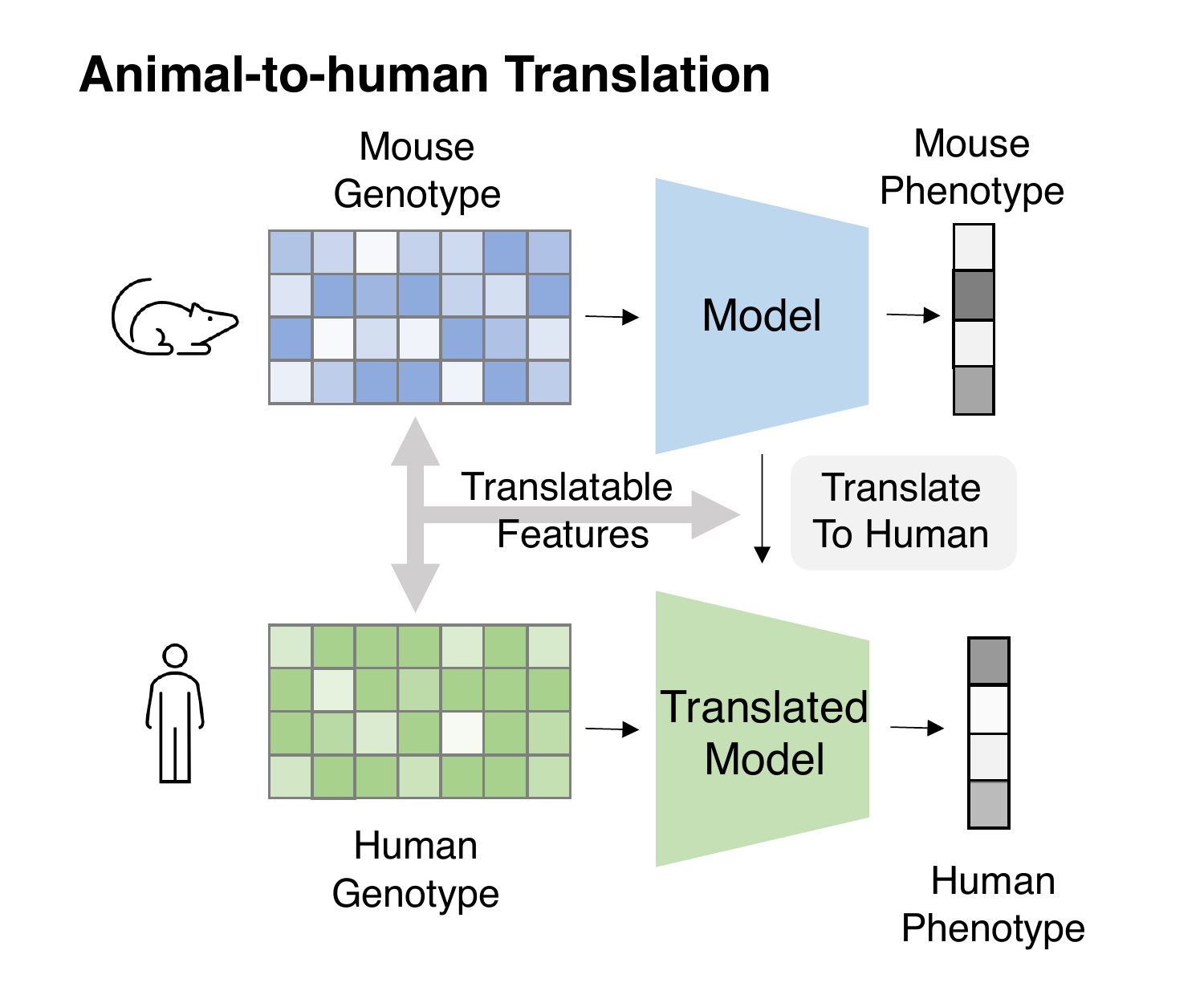}
    \caption{\textbf{Task illustration for the theme "translating preclinical animal models to humans".} A model first obtains translatable features between mouse and human by comparing their genotypes. Next, a predictor model is trained to predict phenotype given mouse genotype. Given the translatable features, augment the predictor and make predictions on human genotypes (Section~\ref{sec:geno-pheno}).}
    \label{fig:translation}
\end{figure}

Before therapeutics move into trials on humans, they are validated through extensive animal model experiments (preclinical studies). However, despite successful preclinical studies,  more than 85\% of early trials for novel drugs fail to translate to humans~\citep{mak2014lost}. One of the main factors for this failure is the gap between animal and human biology and physiology. Animal models do not mimic the human disease condition. However, by comparing large-scale omics data between animals and humans, we can identify translatable features and use machine learning to align animal and human models. 

\subsubsection{Animal-to-human translation} \label{sec:geno-pheno}

One of the central questions of animal-to-human translation is the following. If a study establishes relations between phenotypes and genotypes based on interventions in animals, do these relations persist in humans? Conventional computational methods construct cross-species pairs (CSPs) and compare the pair's molecular profile to find differential expression~\citep{naqvi2019conservation}. Despite identifying several differential features associated with the disease, these methods often do not accurately translate to humans. 

This is where machine learning can help since it is good at making predictions. To formulate it in ML, the genotype-phenotype relations can be captured by some computational model that builds upon an animal's molecular profile (such as using gene expression data to predict disease phenotypes). We can then evaluate the trained computational model to human molecular profiles (test set) and see if the model can accurately predict human phenotypes. A large ML challenge called SBV-IMPROVER was conducted to predict protein phosphorylation on human cells from rat cells using genomics and transcriptomics data under 52 stimulation conditions~\citep{rhrissorrakrai2015understanding}. A wide range of ML approaches such as deep neural networks, trees, and support vector machines were applied and shown to have promising extrapolation performance to humans. 

However, these works directly adopt ML models trained on mice and test on humans, while we know human data often have a different distribution from the mouse data.  This poses a challenge for ML since the ML model often suffers from the out-of-distribution generalizability issue. Recent works have been developed to explicitly model this out-of-distribution property by identifying and leveraging translatable features between animals and humans. \cite{brubaker2019computational} propose a semi-supervised technique that integrates unsupervised modeling of human disease-context datasets into the supervised component that trains on mouse data. In addition, works that directly train on CSPs have also been proposed. For example,  \cite{normand2018found} aim to identify translatable genes. For every gene, they compute the disease effect size for humans and rats in each CSP and apply linear models to fit them. After fitting, they use the mean of the linear model as the predicted human effect size for this gene. They show improved gene selection by up to 50\%. Computational network models leverage existing biological knowledge about system-level signaling pathways and mechanistic models and have shown to identify transferrable biomarkers and predictable pathways~\citep{yao2018integrative,blais2017reconciled}. It is worth noting that the animal-to-human translation problem is similar to the domain adaptation problems in computer vision and natural language processing fields, where they also strive to bridge the gap between the source domain and target domain~\citep{wang2018deep}. Opportunities to leverage advanced domain adaptation techniques to this problem remain open. Despite the improved prediction performance, data availability is still a hurdle to apply ML in this problem since it requires new data for every animal model and disease indication. 

\textit{Machine learning formulations:} Given genotype-phenotype data of animals and only the genotype data of humans, train the model to fit phenotype from the genotype and transfer this model to human. Task illustration is in Figure~\ref{fig:translation}.

\subsection{Curating High-quality Cohorts} \label{sec:cohort}

\begin{figure}
    \centering
    \includegraphics[width=0.9\textwidth]{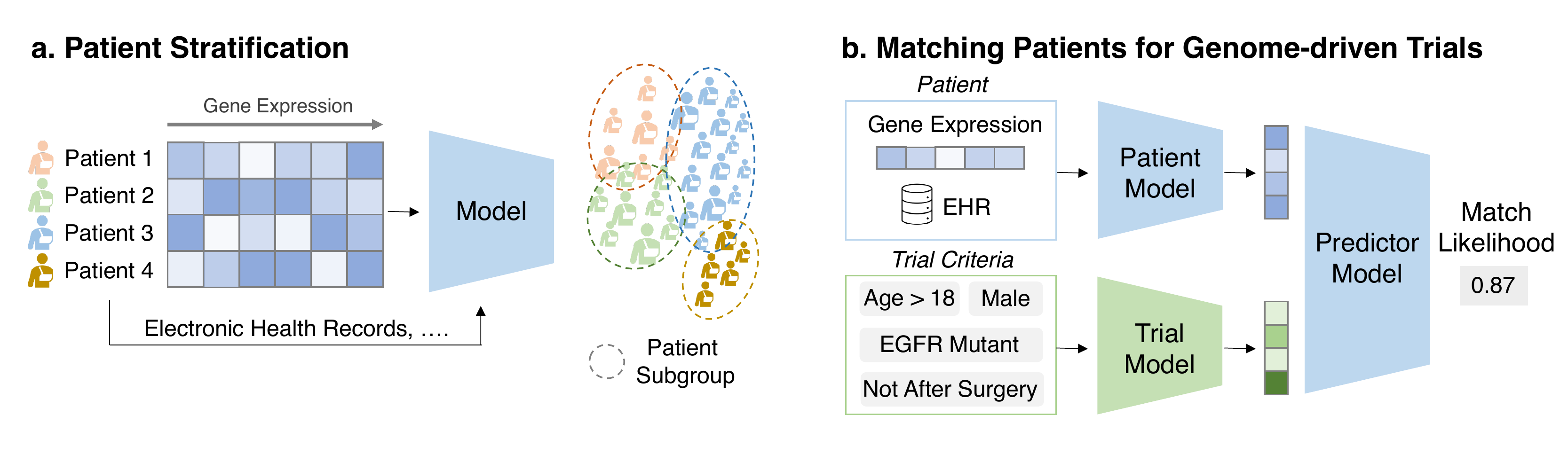}
    \caption{\textbf{Task illustrations for the theme "curating high-quality cohort".} \textbf{a.} Given patient's gene expressions and EHRs, a model clusters them into subgroups (Section~\ref{sec:stratify}). \textbf{b.} A patient model obtains patient embedding from his/her gene expression and EHR. A trial model obtains trial embedding based on trial criteria. A predictor predicts if this patient is fit for enrollment in the given trial (Section~\ref{sec:match}). }
    \label{fig:cohort}
\end{figure}

To study the efficacy of therapeutics in the intended or target patient groups, a clinical trial requires a precise and accurate patient population in each arm~\citep{trusheim2007stratified}. However, due to the heterogeneity of patients, it may be difficult to recruit and enroll appropriate patients. ML can help to characterize important factors for the primary endpoints and quickly identify them in patients by predicting patient molecular profiles. 

\subsubsection{Patient stratification/disease sub-typing}\label{sec:stratify}

Patient stratification in clinical trials is designed to create more homogeneous subgroups with respect to risk of outcome or other important variables that might impact validity of the comparison between treatment arms. Some therapeutics may be highly effective in one patient subgroup, and have a weak or even no effect in other subgroups. In the absence of appropriate stratification in heterogenous patient populations, the average treatment effect across all patients will obscure potentially strong effects in a subpopulation. Conventional stratification methods rely on manual rules on a few available features such as clinical genomics biomarkers, but this might ignore signals rising from rich patient data. Machine learning can potentially identify these important criteria for stratification based on heterogeneous data sources such as genomics profiles, patient demographics, and medical history. 

Various unsupervised models applied on gene expression data have been proposed to group each sample into distinct categories and claim each category as a sub-type. These methods include clustering~\citep{shen2013sparse, witten2010framework}, gene network stratification~\citep{hofree2013network}, and matrix factorization~\citep{gao2005improving}. Also \cite{chen2020deep} proposes a DNN-based clustering method where a supervised constraint on gold-standard sub-type knowledge is included. As the data are high-dimensional and heterogeneous, the challenge is to fuse diverse data sources to obtain a comprehensive patient representation. \cite{wang2014similarity} aggregates mRNA expression, DNA methylation, and microRNA data through similarity network fusion for cancer subtyping. Similarly, \cite{jurmeister2019machine} leverage DNA methylation profiles to subtype lung cancers using DNN and \citep{li2015identification} applies topological data analysis on the patient-patient similarity network constructed from each patient's genotype and EHR data to identify type 2 diabetes subgroups. Despite the accuracy, these methods suffer from interpretability, which is especially important in patient stratification. A black-box stratification model based on complex patient data does not provide a rationale and is often not trustworthy for practitioners to adopt. Decision tree methods are a classical interpretable ML model. Similarly, \cite{valdes2016mediboost} applies a boosted decision tree method with high accuracy compared to a standard decision tree while still providing clues for how the model makes the accurate prediction/stratification. 

\textit{Machine learning formulation: } Given the gene expression and other auxiliary information for a set of patients produce criteria for patient stratification. Task illustration is in Figure~\ref{fig:cohort}a.

\subsubsection{Matching patients for genome-driven trials}\label{sec:match}

Clinical trials suffer from difficulties in recruiting a sufficient number of patients. \cite{mendelsohn2010national} report that 40\% of trials fail to complete accrual in the National Clinical Trial Network and \cite{murthy2004participation} show that less than 2\% of adults with cancer enroll in any clinical trials. Many factors can prevent successful enrollment, such as limited awareness of available trials, and ineffective methods to identify eligible patients in the traditional manual matching system~\citep{lee2019conceptual}. 

These problems can be tackled by automated patient-trial matching, which leverages the heterogeneous patient data such as genomics profile and trial eligibility criteria. Conventional patient-trial matching methods rely on rule-based annotations. For example, \cite{tao2019real} conducts a real-world outcome analysis using an automatic patient-trial matching alert system based on the patient's genomic biomarkers and showed improved results compared to manual matching. However, they are based on heuristics matching rules, which often omits the useful information in rich patient data. \cite{bustos2018learning} uses DNN to generate eligibility criteria, but no matching is done. Recently, advanced machine learning methods have been proposed to leverage the EHR data from patients to match the eligibility criteria of a trial. \cite{zhang2020deepenroll} applies pre-trained Bidirectional Encoder Representations from Transformers(BERT) model for encoding trial protocols into sentence embedding, and uses a hierarchical embedding model to represent patient longitudinal EHR. Building upon this work, \cite{gao2020compose} proposes a multi-granularity memory network to encode structured patient medical codes and use a convolutional highway network to encode trial eligibility criteria. They show significant improvement over previous conventional rule-based methods. However, genomics information has not been included. Methods that fuse genome and EHR data to represent patients could further improve matching efficiency in genome-driven trials. 

\textit{Machine learning formulation: } Given a pair of patient data (genomics, EHR, etc.) and trial eligibility criteria (text description), predict the matching likelihood. Task illustration is in Figure~\ref{fig:cohort}b.

\subsection{Inferring Causal Effects} \label{sec:causal}

\begin{figure}
    \centering
    \includegraphics[width=0.75\textwidth]{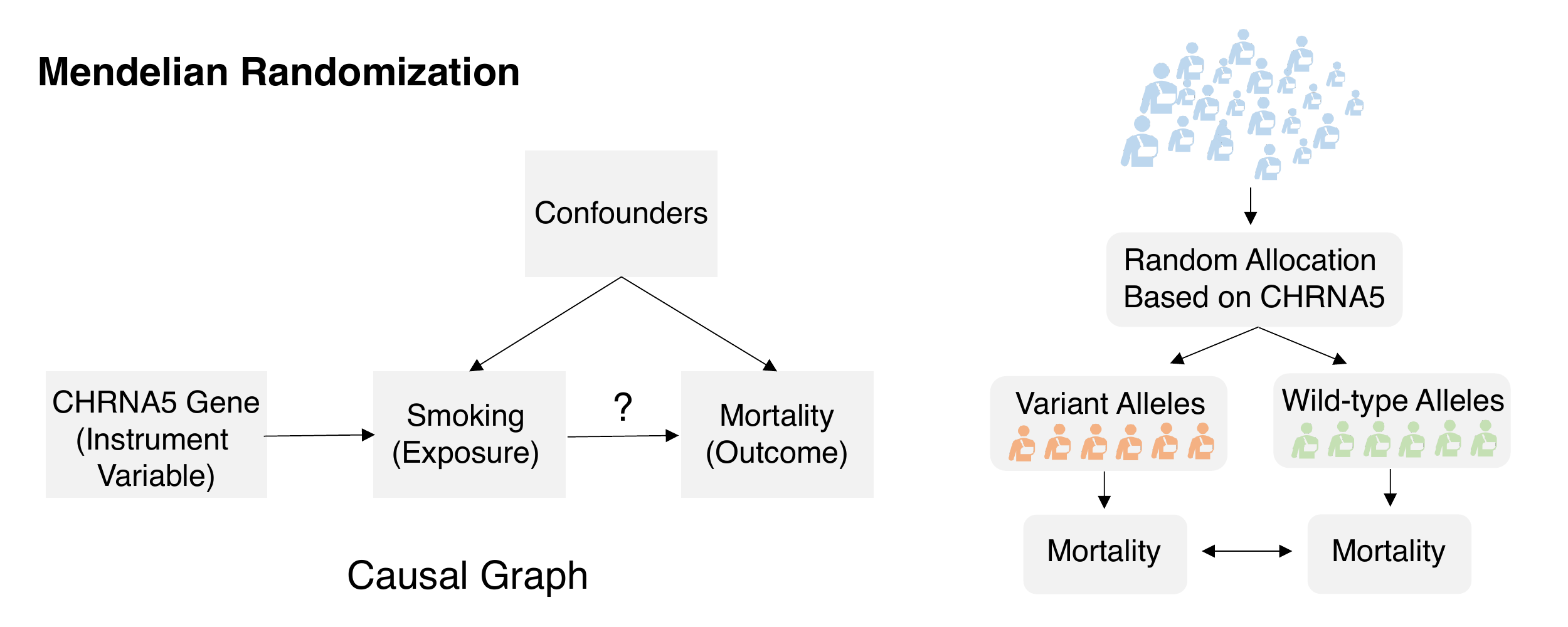}
    \caption{\textbf{Task illustrations for the theme "inferring causal effects".} Left panel: Mendelian randomization relies on using gene biomarker (e.g., CHRNA5) as an instrumental variable to measure the effect of exposure to the outcome as it is not affected by confounders, and it serves as a proxy for exposure by directly comparing the effect of the gene on the outcome. Right panel: patients are first grouped based on the CHRNA5 gene. One group contains variant alleles, and another contains wild-type alleles. Then, the mortality rate can be calculated within each group and compared to see risks. If the risk is high, then we conclude the exposure causes the outcome (Section~\ref{sec:mendelian}). }
    \label{fig:causal}
\end{figure}

Clinical trials study the treatment efficacy on humans. Numerous unmeasured confounders can lead to a biased conclusion about the efficacy. To eliminate these confounders, randomization is conducted such that the control and treatment groups would have an equal distribution of confounders. This way, the comparative effect is not due to unmeasured confounders. However, this requires the control group receives an alternative therapy (e.g. placebo or standard of care). In many studies, it is difficult or unethical to devise and assign placebos/treatments. In these cases, observational studies can be used to study the correlations between exposure (e.g., smoking) to an outcome (e.g., cancer). However, these studies are typically subjected to unmeasured confounding since no randomization is introduced. Recent methods in causal inference provide alternative ways to do randomization through genomics information. 

\subsubsection{Mendelian randomization} \label{sec:mendelian}

Mendelian randomization (MR) uses genes as a mediator for robust causal inference~\citep{davey2003mendelian}. The key is that genetic information is mostly not modified by postnatal events and is thus not susceptible to confounders. If a gene is associated with the exposure and the outcome via the exposure (i.e., vertical pleiotropy), we can use genes as an instrumental variable to simulate randomization. For example, we know that CHRNA5 genes are associated with smoking levels. Then, we can use the CHRNA5 status to group patients and estimate the comparative effect on outcome (e.g., mortality). This process has a tremendous impact as it can bypass clinical trials, add support for trials, and serve as validation for drug targets~\citep{emdin2017mendelian,ference2012effect}. 
Regression analysis is usually conducted to calculate the effects. Despite the promises, challenges remain for more advanced ML and causal inference methods. One challenge is that in some cases, the assumption of vertical pleiotropy does not hold. For example, the genes can associate with the outcome through another pathway (i.e., horizontal pleiotropy)~\citep{verbanck2018detection}. This requires customized probabilistic models and larger sample size for statistically significant estimation~\citep{cho2020exploiting}. The underlying causal pathways among exposures, genes, and outcomes are usually not obvious in many cases due to limited knowledge. A large-scale causal pathway could not only help protect MR from horizontal pleiotropy by knowing when it could be the case but also allows more accurate causal inference with advanced methods by the inclusion of other genes or selection of alternative genes as the instrument variable. The main challenge to obtain this putative causal map is that different models can contradict conclusions given the same dataset. \cite{hemani2017automating} applies a mixture-of-experts random forest framework to reduce the false discovery rate on a set of GWAS data to construct a large-scale causal map of human genome and phenotype and show its usefulness in MR.

\textit{Machine learning formulation: } Given observation data of the genomic factor, exposure, outcome, and other auxiliary information formulate or identify the causal relations among them and compute the effect of the exposure to the outcome. Task illustration is in Figure~\ref{fig:causal}.

\section{Machine Learning for Genomics in Post-Market Studies} \label{sec:post-market}

After a therapeutic is evaluated in clinical trials and approved for marketing, numerous studies are done to monitor its efficacy and safety when used in clinical practice. These studies contain important and often unknown information about therapeutics that was not evident prior to regulatory approval. This section reviews how ML can mine through a large corpus of texts and identify useful signals for post-market surveillance.

\subsection{Mining Real-World Evidence}
\label{sec:rwe}

\begin{figure}
    \centering
    \includegraphics[width=0.95\textwidth]{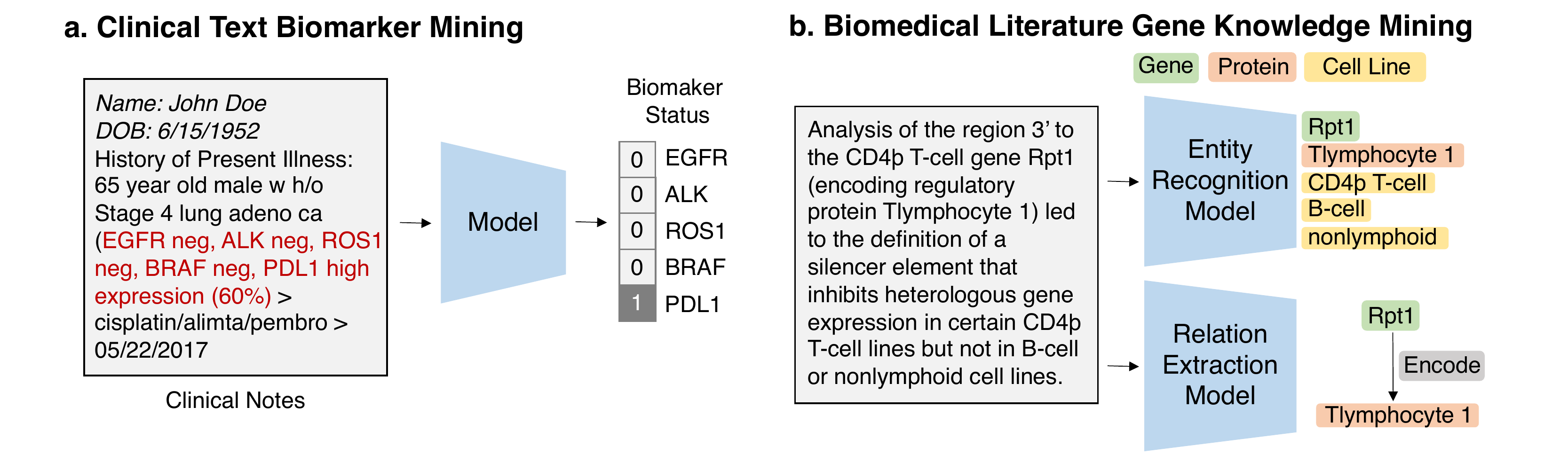}
    \caption{\textbf{Task illustrations for the theme "mining real-world evidence".} \textbf{a.} A model predicts genomic biomarker status given a patient's clinical notes (Section~\ref{sec:clinical_text}). \textbf{b.} A model recognizes entities in the literature and extracts relations among these entities (Section~\ref{sec:biomed_literature}). Credits: the text in panel a is from~\cite{huang2020interpretable}; the text in panel b is from~\cite{zhu2018gram}.}
    \label{fig:rwe}
\end{figure}

After therapeutics are approved and used to treat patients, voluminous documentation is generated in the EHR system, insurance billing system, and scientific literature. These are called real-world data. The analyses of these data are called real world evidence. They contain important insights about therapeutics, such as patients' drug responses given different patient characteristics. They can also shed light on disease mechanism of action, the novel phenotype for a target gene, and so on. However, free-form texts are notoriously hard to process. Natural language processing (NLP) technology can be helpful to mine insights from these texts. Next, we describe two specific tasks involving two types of real-world evidence, namely, clinical notes and scientific literature. 

\subsubsection{Clinical Text Biomarker Mining} \label{sec:clinical_text}

EHR has rich information about the patient, and it records a wide range of patient's vitals and disease course after treatments. This information is critical for post-market research, where an actionable hypothesis can be drawn. However, the structured EHR data does not cover the entire picture of a patient. The majority of important variables can only be found in the clinical notes~\citep{boag2018s}, such as next-generation sequencing (NGS) status, PDL1 (Immunotherapy) status, treatment change, and so on. These variables can directly facilitate predictive model building to support clinical decision-making or increase the power of disease-gene-drug associations to better understand the drug. However, conventional human annotations are costly, time-consuming, and not scalable. 

Automatic processing of clinical notes of patients using machine learning can facilitate this process. For example, \cite{guan2019natural} uses bidirectional LSTMs to extract NGS-related information in a patient's genetics report and classify documents to the treatment-change and no-treatment-change groups. However, clinical text is very messy, filled with typos and jargon (e.g., acronyms). Standard NLP techniques do not work. Also, clinical text often requires clinical annotations. Specialized machine learning models are required, such as transfer learning techniques that learn a sufficient clinical note representation through large-scaled self-supervised learning on clinical notes and fine-tuning on a task of interest with a small number of annotations~\citep{devlin2018bert,huang2019clinicalbert}. \cite{huang2020interpretable} applies hierarchical BERT-based models to classify PDL1 and NGS status and use an attention mechanism to provide clues for which parts of a text provide these variables. 

\textit{Machine learning formulation: } Given a clinical note document, predict the genomic biomarker variable of interest. Task illustration is in Figure~\ref{fig:rwe}a.

\subsubsection{Biomedical Literature Gene Knowledge Mining}\label{sec:biomed_literature}

One key question in post-market research is to find evidence about a therapeutics’ response to diseases given patient characteristics such as genomic biomarkers. This has several important applications such as validation of therapeutic efficacy, identification of potential off-label genes/diseases for drug repurposing, and detection of therapeutic candidates’ adverse events when treating patients, using some genomic biomarkers. They also serve as important complementary information for target discovery. This summarized information about drug-gene and disease-gene relations is usually reported and published in the scientific literature. Manual annotations are infeasible due to the exponential number of new articles published every day. 

Conventional methods are rule-based~\citep{tsai2006nerbio} and dictionary-based~\citep{hirschman2002rutabaga}. They both rely on hand-crafted rules/features to construct query text templates and search through the papers to find sentences that match these templates~\citep{davis2013ctd}. However, these hand-crafted features require extensive domain knowledge and are difficult to keep up-to-date with new literature. The limited flexibility leads to the omission of potential newly discovered drug-gene/drug-disease pairs. Recent advances in name entity recognition and relation detection through deep learning can automatically learn from a large corpus to obtain an optimal set of features without human engineering and have shown strong performances~\citep{nasar2018information}. This can be formulated as a model to recognize drugs, genes, disease terms, and detect drug-gene or drug-disease relation types given a set of documents. Numerous machine learning methods have been developed for biomedical named entity recognition/relation extraction. For example, \cite{limsopatham2016learning} uses bidirectional LSTM to predict the name entity label for each word with character-level embedding. \cite{zhu2018gram} use an n-gram based CNN to capture local context around each word for improved prediction. 

On relation extraction, in addition to the CNN~\citep{zhao2016drug} and RNN~\citep{zhang2018drug} architecture, \cite{zhang2018hybrid} proposes a hybrid model that integrates a CNN on syntax dependency tree and an RNN on the sentence encodings for improved biomedical relation prediction. \cite{zhang2018graph} applies a graph convolutional neural network on the syntax dependency tree of a sentence and shows improved relation extraction.  ML models require large amounts of label annotations as training data, which can be difficult to obtain. Distant supervision borrows information from a large-scale knowledge base to automatically create labels so that it does not require labeled corpora, which reduces manual annotation efforts. \cite{lamurias2017extracting} applies a distant-learning based pipeline that predicts microRNA-gene relations. Recently, BioBERT extends BERT~\citep{devlin2018bert} to pre-train on a large-scale biomedical scientific literature corpus and fine-tune it on numerous downstream tasks and has shown great performance in benchmarking tasks such as biomedical named entity recognition and relation extraction. 

\textit{Machine learning formulation: } Given a document from literature, extract the drug-gene, drug-disease terms, and predict the interaction types from the text. Task illustration is in Figure~\ref{fig:rwe}b.

\section{Discussion: Open Challenges and Opportunities} \label{sec:challenge}

This survey provides an overview of research in the intersection of machine learning, genomics, and therapeutic developments. It is our view that machine learning has the potential to revolutionize the use of genomics in therapeutics development, as we have presented a diverse set of such applications in Sections~\ref{sec:target}-\ref{sec:post-market}. However, numerous challenges remain. Here, we discuss these challenges and the associated opportunities. 

\xhdr{Distribution shifts}
ML models work well when the training and deployment data follow the same data distribution. However, in real-world usage in genomics and therapeutics ML, many problems experience distribution shifts where the deployment environment and the data generated from it are different from the training stage. For example, training happens given the available batches of gene expression data in brain tissue. The resulting model is required to predict a new experiment with bone tissue. Another example is to train on animal model transcriptomics and predict the phenotype of human models. Thus, a model must generalize to out-of-distribution. Distribution shifts have been a longstanding challenge in ML, and a large body of work in model robustness and domain adaptation could be applied to genomics to improve generalizability~\citep{moreno2012unifying}. For instance, \cite{brbic2020mars} utilizes the meta-learning technique to generalize to novel single-cell experiments.  

\xhdr{Learning from small datasets} Biological data are generated through expensive experiments. This means that many tasks only have a minimal number of labeled data points. For example, there are usually only a few drug response data points for new therapeutics. However, standard ML models, especially DL models, are data-hungry. Thus, how to make an ML model learn given only a few examples is crucial. Transfer learning can learn from a large body of existing labeled data points and transfer it to the downstream task with limited data points~\citep{devlin2018bert}. However, they usually still require a reasonable number of training data points.  Given only a few data points, few-shot learning methods such as model-agnostic meta-learning (MAML)~\citep{finn2017model}, prototypical networks~\citep{snell2017prototypical} learning from other related tasks using a few examples have shown strong promises. Recently, \cite{ma2021few} have successfully applied MAML to improve few-shot drug response prediction.

\xhdr{Representation capacity} 
The key to successful ML models depends on the effective representation of the genome and other related biomedical entities that match biological motivation. For example, the current dominant ML model for DNA sequence is through CNN models. However, most successful usage only applies to short DNA sequences generated from predefined preprocessing steps, instead of a large fraction of the whole-genome sequence, which could allow a model to tap into crucial information of long-range gene regulatory dependencies~\citep{alipanahi2015predicting,deepsea}. RNN and transformers are also only able to take in medium-length inputs, in contrast to more than $O(10^6)$ SNPs per genome. This also means that the number of input features can be orders of magnitude larger than the number of data points, a well-known ML challenge called the curse of dimensionality. Furthermore, the general ML models are often developed for image and text data without any biological motivations. Thus, to model the human genome and the complicated regulation among genes, a domain-motivated model that captures interactions among extremely long-range high-dimensional features is needed. Initial attempts for domain-motivated representation learning have been made. For instance, \cite{romero2016diet} proposes a parameter prediction network that reduces the number of free parameters for DNN to alleviate the curse of dimensionality issue and shows improved patient stratification given $10^6$ SNPs. \cite{ma2018using} modifies the neural network structure to simulate the hierarchical biological processes and explain pathways for phenotype prediction. 

\xhdr{Model trustworthiness} For an ML model to be used by domain scientists, the model prediction has to be trustworthy. This can happen on two levels. First, in addition to accurate prediction, the model prediction also needs to generate justification in terms of biomedical knowledge ({\it explanation}). However, current ML models focus on improving model prediction accuracy. Towards the goal of explanation, ML models need to encode biomedical knowledge. This can be potentially achieved via integrating biological knowledge graphs~\citep{himmelstein2017systematic} and applies the graph explainability method~\citep{ying2019gnnexplainer}. 

The second level is on the quality of model prediction. Since ML models are not error-free, it is important to alert the users or abstain from making predictions when the model is not confident. Uncertainty quantification or model abstention around the model prediction can alleviate this problem. Recently, \cite{hie2020leveraging} uses Gaussian processes to generate uncertainty scores of compound bioactivity, protein fluorescence, and single-cell transcriptomic imputation and has been shown to guide the experimental and validation loop. Integrating the explanation into human workflows and promoting human trust in AI also requires special attention as recent works show that directly providing AI explanation to humans can confuse the human observer and degrade the performance~\citep{bansal2020does}.

\xhdr{Fairness} ML models can manifest the bias in the training data. It has been shown that ML models do not work equally well on all subpopulations. These algorithmic biases could have significant social and ethical consequences. For example, \cite{martin2018hidden} found that $79\%$ of genomic data are from patients of European descent, even though they comprise only $16\%$ of the world’s population. Due to differences in allele frequencies and effect sizes across populations, ML models that perform well on the discovery population generally have much lower accuracy and are worse predictors in other populations. Since most discovery to date is performed with European-ancestry cohorts, predictive models may exacerbate health disparities since they will not be available for or have lower utility in African and Hispanic ancestry populations. Similarly, most studies focus on common diseases, whereas experimental data on rare diseases are often limited. These imbalances against minorities require specialized ML techniques. The fairness in ML is defined to make the prediction independent of protected variables such as race, gender, and sexual orientation~\citep{barocas2017fairness}. Recent works have been proposed to ensure this criterion in the clinical ML domain~\citep{pierson2021algorithmic}. However, fairness research of ML for the genomic domain is still lacking.

\xhdr{Data linking and integration} An individual has a diverse set of data modalities, such as genomics, transcriptomics, proteomics, electronic health records, and social-economic data. Current ML approaches focus on developing methods for a single data modality, whereas to fully capture the comprehensive data types around individuals could potentially unlock new biological insights and actionable hypotheses. One of the reasons for the limited integration is the lack of data access that connects these heterogeneous data types. As large-scale efforts such as UK Biobank~\citep{canela2018atlas}, which connects in-depth genetic and EHR information about a patient, become available, new ML methods designed to take into account this heterogeneity are needed. Indeed, recent studies have discovered novel insights by applying ML to linked genomics and EHR data~\citep{shen2019brain,willetts2018statistical,bellot2018can}. 
Handling missing data across modalities is a common challenge in this setting.

\xhdr{Genomics data privacy} Abundant genomics data and annotations are generated every day. Aggregation of these data and annotations can tremendously benefit ML models. However, these are usually considered private assets for individuals and contain sensitive private information and thus are not shareable directly. Techniques to anonymize, de-identify these data using differential privacy can potentially enable genomics data sharing~\citep{azencott2018machine}. In addition, recent advances in federated learning techniques allow machine learning model training on aggregated data without sharing data~\citep{yang2019federated}.


\section{Conclusion} \label{sec:conclusion}
We have conducted a comprehensive review of the literature on machine learning applications for genomics in therapeutics development. We systematically identify diverse ML applications in genomics and provide pointers to the latest methods and resources. For ML researchers, we show that most of these applications have problems that remain unsolved and thus provide numerous exciting technical challenges for ML method innovations. We also provide concise ML problem formulation to help ML researchers to approach these tasks. For biomedical researchers, we pinpoint a large set of diverse use cases of ML applications, where they can extend and expand novel use cases. We also introduced the popular ML models and their corresponding use cases in genomic data. 

In conclusion, this survey provides an in-depth research summary of the intersection of machine learning, genomics, and therapeutic developments. We hope this survey can lead to a deeper understanding of this interdisciplinary domain between ML and genomics and broaden the collaboration across these two communities. As a common belief that the future of medicine is personalized, understanding the therapeutic tasks with machine learning methods on genomic data is the key to lead ultimate breakthroughs in drug discovery and development. We hope this survey helped to bridge the gap between genomics and machine learning domains.



\bibliographystyle{agsm}
\bibliography{ref}

\clearpage
\appendix

\end{document}